\documentclass[preprint]{elsarticle}
\usepackage{amsmath, amssymb, amsxtra, amsfonts}
\usepackage{graphicx,paralist}
\usepackage{url}
\usepackage{tikz}
\usetikzlibrary{trees}
\usepackage{xspace}
\usepackage{hyperref}
\usepackage{setspace}

\newtheorem{theorem}{Theorem}[section]
\newtheorem{corollary}[theorem]{Corollary}

\newtheorem{proposition}{Proposition}
\newtheorem{definition}{Definition}

\newtheorem{example}[theorem]{Example}

\def\0{{\mathbf 0}}
\def\1{{\mathbf 1}}
\def\C{{\mathcal C}}

\def\0{{\mathbf 0}}
\def\1{{\mathbf 1}}
\def\C{{\mathcal C}}
\long\def\comment#1{}

\begin{document}
\begin{frontmatter}

\title{ConArg: a Tool to Solve (Weighted) Abstract  Argumentation Frameworks with (Soft) Constraints}

\author[perugia]{Stefano Bistarelli}
\ead{bista@dmi.unipg.it}
\author[inria]{Francesco Santini\corref{cor1}}
\ead{francesco.santini@inria.fr}

\address[perugia]{Dipartimento di Matematica e Informatica, Universit\`a di Perugia, Italy}
\address[inria]{Contraintes, INRIA-Rocquencourt, France}

\cortext[cor1]{Corresponding author}

\begin{abstract} 
\emph{ConArg} is a  \emph{Constraint Programming}-based tool  that can be used to model and solve
different problems related to  \emph{Abstract Argumentation Frameworks} (AFs). To implement this tool we have used \emph{JaCoP}, a Java library that provides the user with a \emph{Finite
Domain Constraint
Programming} paradigm. ConArg is  able to randomly generate networks with small-world properties in order to find conflict-free, admissible, complete, stable grounded, preferred, semi-stable, stage and ideal extensions on such interaction graphs. We present the main features of ConArg and we report the performance in time, showing also a comparison with ASPARTIX~\cite{aspartix}, a similar tool using \emph{Answer Set Programming}. The use of techniques for constraint solving can tackle the complexity of 
the problems presented in~\cite{main2}. Moreover we suggest semiring-based soft constraints as a mean
to parametrically represent and solve  \emph{Weighted Argumentation
Frameworks}: different kinds of preference levels related to
attacks, e.g., a score representing a ``fuzziness'', a ``cost''
or a probability, can be represented by
choosing different  instantiation of the semiring algebraic structure. The basic idea
is to provide a common computational and quantitative framework.  
\end{abstract}


\end{frontmatter}


\section{Introduction}\label{sec:Introduction}

\emph{Argumentation}~\cite{bookargumentation} is based on the exchange and the
evaluation of interacting arguments which may represent
information of various kinds, especially beliefs or goals.
Argumentation  can be used for modeling some aspects of reasoning,
decision making, and dialogue. For instance, when
an agent has conflicting beliefs (viewed as arguments), a
(nontrivial) set of plausible consequences can be derived through
argumentation from
the most acceptable arguments for the agent. Argumentation has become an important subject of research in
Artificial Intelligence and it is also of interest in several
disciplines, such as Logic, Philosophy and Communication
Theory~\cite{bookargumentation,mogdil}.

Many  theoretical and practical developments build on Dung's
seminal theory of argumentation. A \emph{Dung's Abstract Argumentation
Framework}  (\emph{AF}) or \emph{Abstract Argument System} is a directed graph consisting of a set of
arguments and a binary conflict based \emph{attack relation} among
them~\cite{dung,bookargumentation}. The sets of arguments to be considered are then defined
under different semantics, where the choice of semantics equates
with varying
degrees of scepticism or credulousness.  The main issue for
any theory of argumentation is the selection of acceptable sets of
arguments, based on the way arguments interact. Intuitively, an
acceptable set of arguments must be in some sense coherent and
strong enough (e.g., able to defend itself against all attacking
arguments).

In this paper we propose \emph{ConArg} (i.e., ``Argumentation with Constraints''), a Java-based tool that can find all the  classical extensions proposed by Dung~\cite{dung}, i.e., conflict-free, admissible, complete, stable, preferred and grounded, other successively ideated extension as semi-stable~\cite{semistable}, stage~\cite{stage} and ideal~\cite{ideal}, and it can also solve the hard problems related to \emph{Weighted Argumentation Frameworks} (\emph{WAF}),  which have been presented in \cite{main,main2}. An example of these hard problems that ConArg is able to solve, is, given a WAF, a set of arguments and \emph{inconsistency budget} $\beta$~\cite{main,main2}, checking whether $\beta$ is minimal or not. This specific problem is co-NP-complete~\cite{main,main2}. 

As the core of our solver we decide to use \emph{Constraint Programming} (CP)~\cite{bookrossi}, which is  a powerful paradigm for solving combinatorial search problems that draws on a wide range of techniques from artificial intelligence, computer science, databases, programming languages, and operations research. Constraint programming is currently applied with success to many domains, such as scheduling, planning, vehicle routing, configuration, networks, and bioinformatics~\cite{bookrossi}. 
Constraint solvers search the solution space either systematically, as with backtracking or branch and bound algorithms, or use forms of local search that may be incomplete. An instance of a \emph{Constraint Satisfaction Problem} (\emph{CSP})~\cite{bookrossi}, as formally presented in Section~\ref{sec:softbg}, describes a problem in terms of constraints.

To solve problems related to WAFs we use semiring-based \emph{Soft Constraint Programming}~\cite{jacm97,bistabook}  instead. The key
idea behind this formalism is to extend the classical notion of constraint  by adding a structure representing its level of satisfiability (or preference/cost), that is a semiring-like structure (see Section~\ref{sec:soft}).

Even finding all the classical Dung's extensions is not ``easy'': the number of these extensions, which in practice are 
subsets of the $\mathcal{A}_{rgs}$ set of arguments, may explode for large $\mathcal{A}_{rgs}$ (the powerset of  $\mathcal{A}_{rgs}$ has $2^{|\mathcal{A}_{rgs}|}$ elements). Therefore, it is important to use techniques to tackle this inherent complexity, as those ones adopted in CP. This is particularly important with conflict-free extensions, which represent the ``least constrained'' extensions. 

To model all the introduced  problems with constraints, we adopt  \emph{Java Constraint Programming} solver\footnote{\url{http://www.jacop.eu}} (JaCoP), a Java library that provides the Java user with a \emph{Finite Domain Constraint Programming} paradigm~\cite{bookrossi}.  With ConArg, the user can import  an interaction graph as a textual description file, or he can generate the input according to two different kinds of small-world networks: \emph{Barabasi}~\cite{Barabasi} and \emph{Kleinberg}~\cite{kleinberg} graphs. We suppose that interaction graphs, where nodes are
arguments and edges are attacks (see Section~\ref{sec:argurelated}),
represent in this case a kind of  social network, and consequently show the
related small-world properties~\cite{discussion,facebook}. A practical example can be the study of discussion fora, where the users post their arguments that can attack other users' arguments~\cite{facebook,socialarg}.

This work details, integrates, and extends with a description of the ConArg tool the research line previously proposed in  \cite{ecai10,tafa11,ictai11}. The remainder of this paper is organized as follows. In
Section~\ref{sec:argurelated}  we report the theory behind AFs
and, in Section~\ref{sec:bgweightedAF}, about the WAF formalism presented in \cite{main,main2}. In Section~\ref{sec:softbg} we summarize the
background on CP~\cite{bookrossi} and on its soft extension proposed in \cite{jacm97,bistabook}. In
Section~\ref{mapping} we show the mapping from AFs to CSPs, which is at the heart of ConArg: by solving the CSP, we find a solution of the related AFs (e.g., all the conflict-free extensions). 

Section~\ref{sec:argsemi} revises the unifying WAF formalism we originally proposed in \cite{ecai10}, which is based on the notion of semiring structure~\cite{jacm97,bistabook}. Afterwards, in Section~\ref{sec:mapping2} we show how we have implemented in ConArg the two WAFs respectively presented in Section~\ref{sec:argsemi}, and, in Section~\ref{sec:bgrounded}, the WAF advanced in \cite{main,main2} and briefly reported in Sec.~\ref{sec:bgweightedAF}.

In Section~\ref{ConArg} we describe the main features of ConArg by also showing some screenshots of the the application we developed, while in Section~\ref{sec:tests} we  report the performance in time of our constraint-based search; in Section \ref{sec:aspartix} we also show a performance comparison between our solution and the \emph{ASPARTIX} system~\cite{aspartix}. A comparison with related work is given in
Section~\ref{sec:related} instead. Finally, Section~\ref{sec:conclusions}
draws the conclusive remarks and outlines future work. 

\section{Background on Argument Systems}
\label{sec:argurelated} In~\cite{dung}, the author has proposed an
abstract framework for argumentation in which he focuses on the
definition of the status of arguments. For that purpose, it can be
assumed that a set of arguments is given, as well as the different
conflicts among them. An argument is an abstract entity whose role
is solely determined by its relations to other arguments.

\begin{definition}\label{def1}
An Argumentation Framework (AF) is a pair $\langle
\mathcal{A}_{rgs},R \rangle$ of a set $\mathcal{A}_{rgs}$ of
arguments and a binary relation $R$ on $\mathcal{A}_{rgs}$ called
the attack relation. $\forall a_i, a_j \in \mathcal{A}_{rgs}$, $a_i R\, a_j$ means
that $a_i$ attacks $a_j$. An AF may be represented by a directed
graph (the interaction graph) whose nodes are arguments and edges
represent the attack relation. A set of arguments $\mathcal{B}$
attacks an argument $a$ if $a$ is attacked by an argument of
$\mathcal{B}$. A set of arguments $\mathcal{B}$ attacks a set of
arguments $\mathcal{C}$ if there is an argument $b \in
\mathcal{B}$ which attacks an argument $c \in \mathcal{C}$.
\end{definition}

 The ``acceptability'' of
an argument~\cite{dung} depends on its
membership to some sets, called \emph{extensions}. These
extensions characterize collective ``acceptability''. 

\begin{figure}[h]
\centering
\includegraphics[scale=0.52]{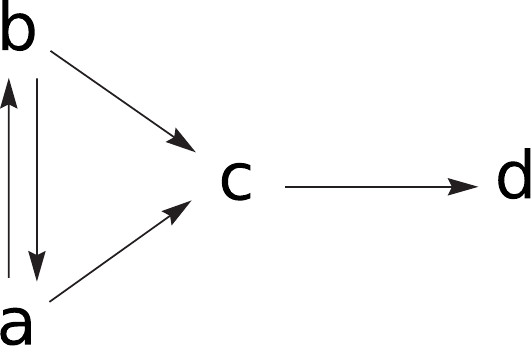} 
\caption{An example of Dung Argumentation Framework; e.g., $c$
attacks $d$.} \label{figure:argexample}
\end{figure}

In Figure~\ref{figure:argexample} we show an example of AF
represented as an \emph{interaction graph}: the nodes represent
the arguments and the directed arrow from $c$ to $d$ represents
the attack of $c$ towards $d$, that is $c \, R\, d$.
Dung~\cite{dung} gave several semantics to ``acceptability''. These
various semantics produce none, one or several acceptable sets of
arguments, called extensions. In Def.~\ref{def2e} we define the concepts of conflict-free and stable extensions:

\begin{definition}\label{def2e}
A set $\mathcal{B} \subseteq \mathcal{A}_{rgs}$ is conflict-free
iff no two arguments $a$ and $b$ in $\mathcal{B}$ exist
such that $a$ attacks $b$. A conflict-free set $\mathcal{B}
\subseteq \mathcal{A}_{rgs}$ is a stable extension iff for each
argument which is not in $\mathcal{B}$, there exists an argument
in $\mathcal{B}$ that attacks it.
\end{definition}

The other semantics for ``acceptability'' rely upon the concept of
defense:

\begin{definition}\label{def2}
An argument $b$ is defended by a set $\mathcal{B} \subseteq
\mathcal{A}_{rgs}$ (or $\mathcal{B}$ defends $b$) iff for any
argument $a \in \mathcal{A}_{rgs}$, if $a$ attacks $b$ then
$\mathcal{B}$ attacks $a$.
\end{definition}

An admissible set of arguments according to Dung must be a
conflict-free set which defends all its elements. Formally:

\begin{definition}\label{def3}
A conflict-free set $\mathcal{B} \subseteq \mathcal{A}_{rgs}$ is
admissible iff each argument in $\mathcal{B}$ is defended by
$\mathcal{B}$.
\end{definition}

Besides the stable semantics, three semantics refining
admissibility have been introduced by Dung~\cite{dung}:

\begin{definition}\label{def4}
A preferred extension is a maximal (w.r.t. set inclusion)
admissible subset of $\mathcal{A}_{rgs}$. An admissible
$\mathcal{B} \subseteq \mathcal{A}_{rgs}$ is a complete extension
iff each argument which is defended by $\mathcal{B}$ is in
$\mathcal{B}$. The least (w.r.t. set inclusion) complete extension
is the grounded extension.
\end{definition}

A stable extension is also a preferred extension and a preferred
extension is also a complete extension. Stable, preferred and
complete semantics admit multiple extensions whereas the grounded
semantics ascribes a single extension to a given argument system.

The definitions of stage~\cite{stage} and semi-stable~\cite{semistable} semantics are based on the idea of prescribing the maximization not only of the arguments included in the extension (as for the preferred extension) in Def.~\ref{def4}, but also of those attacked by it:

\begin{definition}\label{def5}
Given a set $\mathcal{B} \subseteq \mathcal{A}_{rgs}$, the \emph{range} of $\mathcal{B}$ is defined as $\mathcal{B} \cup \mathcal{B}^{+}$, where $\mathcal{B}^{+}= \{a \in \mathcal{A}_{rgs} : \mathcal{B} \text{ attacks } a\}$. $\mathcal{B}$ is a stage extension iff $\mathcal{B}$ is a conflict-free set with maximal (w.r.t. set inclusion) range. $\mathcal{B}$ is a semi-stable extension iff $\mathcal{B}$ is a complete extension  with maximal (w.r.t. set inclusion) range.
\end{definition}

Ideal semantics~\cite{ideal}, defined in Def.~\ref{def6}, provides a unique-status approach allowing the acceptance of a set of arguments possibly larger than in the case of the grounded extension.

\begin{definition}\label{def6}
A set $\mathcal{B} \subseteq \mathcal{A}_{rgs}$ is ideal iff $\mathcal{B}$ is admissible and for each preferred extensions $\mathcal{E}$, then $\mathcal{B} \subseteq \mathcal{E}$. The ideal extension is the maximal (w.r.t. set inclusion) ideal set.
\end{definition}

\subsection{Weighted Argumentation Frameworks and Related Hard Problems}\label{sec:bgweightedAF}
In ConArg we also solve  hard problems related to WAFs~\cite{main,main2}. Formally,  a WAF	is a triple $\langle
\mathcal{A}_{rgs}, R, w \rangle$ where $\langle
\mathcal{A}_{rgs}, R \rangle$ is a Dung-style abstract argument system, and $w : R \rightarrow \mathbb{R}^+$ is a function assigning real valued weights  to attacks. 

A key idea presented in~\cite{main,main2} is the \emph{inconsistency budget}, $\beta \in \mathbb{R}^+$, which the authors use to characterise how much inconsistency they are prepared to tolerate. The intended interpretation is that, given an inconsistency budget $\beta$, \emph{we would be prepared to disregard attacks up to a total weight of }$\beta$~\cite{main,main2}. Conventional AFs implicitly assume an inconsistency budget of $0$. In Section~\ref{sec:argsemi} we focus on WAFs either, by considering a semiring-based constraint programming framework: the solution of these representations is implemented in ConArg as well.

 As shown in~\cite{main,main2}, while the the problem of finding the weighted version of the classical extensions (e.g., stable or admissible) is not computationally harder than the original problem, there are some important problems related to weighted grounded extensions  that are very difficult to solve. The concept of inconsistency budget $\beta$ has been introduced in Section~\ref{sec:argurelated}. 
 
In the following propositions, i.e., Proposition~\ref{prop1}, Proposition~\ref{prop2} and Proposition~\ref{prop3}, we show three complex problems proposed in~\cite{main,main2}. As for preferred extensions, we say an argument is credulously accepted if it forms a member of at least one weighted grounded extension, and sceptically accepted if it is a member of every weighted grounded extensions~\cite{main,main2}. Since there are multiple $\beta$-grounded extensions~\cite{main,main2}, we can consider credulous and sceptical variations of the problem, as with preferred extensions. In Proposition~\ref{prop1} we consider the credulous case first:

\begin{proposition}[\cite{main,main2}]\label{prop1} Given a weighted argument system $\langle \mathcal{A}_{rgs}, R, w \rangle$, an inconsistency budget $\beta$, and argument $a \in \mathcal{A}_{rgs}$, the problem of checking whether $\exists L \in wge(\mathcal{A}_{rgs}, R, w, \beta)$ such that $a \in L$ is NP-complete. 
\end{proposition}

In Proposition~\ref{prop2}  we consider the  ``sceptical'' version of the problem.

\begin{proposition}[\cite{main,main2}]\label{prop2}Given a weighted  argument system  $\langle \mathcal{A}_{rgs}, R, w \rangle$, an inconsistency budget $\beta$, and an argument $a \in \mathcal{A}_{rgs}$, the problem of checking whether, $\forall L	\in wge(\mathcal{A}_{rgs}, R, w, \beta)$,	we	have	 $a \in L$	is co-NP-complete.
\end{proposition}

Suppose now we have a weighted argument system $\langle X, A, w \rangle$ and a set of arguments $S$. Then, what is the smallest amount of inconsistency would we need to tolerate in order to make $S$ a solution? When considering conflict-free and admissible extensions, the answer is easy: we know exactly which attacks we would have to disregard to make a set of arguments admissible or consistent. However, when considering weighted grounded extensions, the answer is not so easy. There may be multiple ways of getting a set of arguments into a weighted extension, each with potentially different costs; we are thus typically interested in solving the problem expressed by Proposition~\ref{prop3}:

\begin{proposition}[\cite{main,main2}]\label{prop3}  Given a weighted argument system  $\langle \mathcal{A}_{rgs}, R, w \rangle$, a set of arguments $L \subseteq \mathcal{A}_{rgs}$, and an inconsistency budget $\beta$, checking whether $\beta$ is minimal w.r.t. $\langle \mathcal{A}_{rgs}, R, w \rangle$ and $L$ is \emph{co-NP-complete}.
\end{proposition}

\section{Constraint Programming}
\label{sec:softbg}

A \emph{Constraint Satisfaction Problem} (\emph{CSP})~\cite{bookrossi} is defined as a triple $P = \langle V,D,C \rangle$, where $X$ is set of 
variables $V = \{x_1,x_2, \dots, x_n\}$, \emph{D} is a set 
of domains $D = \{ D_1, D_2, \dots, D_n \}$ such that $x_i \in D_i$, $C$ is a 
set of constraints $C = \{c_1, c_2, \dots , c_t \}$. A constraint $c_j$ 
is a pair $\langle R_{O_j} , O_j \rangle$ where $R_{O_j}$ is a relation on the variables 
in $O_j= \textit{scope}(c_j)$. In other words, $R_i$ is a subset of the Cartesian product of 
the domains of the variables in $O_j$.
A solution to the CSP $P$ is an $n$-tuple $T = \langle t_1, t_2, \dots , t_n \rangle$ 
where $t_i \in D_i$ and each $c_j$ is satisfied in that $R_{O_j}$ holds on the 
projection of $T$ onto the scope $O_j$. In a given task one may be required to 
find the set of all solutions, $Sol(P)$, to determine if that set is non-empty or just to 
find any solution, if one exists. If the set of solutions is empty the CSP is unsatisfiable. 
This simple but powerful framework captures a wide range of significant applications 
in fields as diverse as artificial intelligence, operations research, scheduling, supply 
chain management, graph algorithms, computer vision and computational linguistics~\cite{bookrossi}.

One of the main reasons why constraint programming quickly found its way into applications has been the early availability of usable constraint programming systems, as JaCoP, which we will use in the implementation and solution of the AFs. Various generalizations of the classic CSP model have been developed subsequently. One of the most significant is the \emph{Constraint Optimization Problem} (\emph{COP}) for which there are several significantly different formulations, and the nomenclature is not always consistent~\cite{bookrossi}. Perhaps the simplest COP formulation retains the CSP limitation of allowing only ÔhardÕ Boolean-valued constraints but adds a cost function over the variables, that must be minimized. A \emph{weighted constraint} $\langle c, w \rangle$ is just a classical constraint $c$, plus a weight $w$ (over natural, integer, or real numbers).
The cost of an assignment $t$ of the variable is the sum of all $w(c)$, for all constraints $c$ which are violated by $t$~\cite{bookrossi}.

Then, the overall degree of satisfaction (or violation) of the assignment is obtained by combining these elementary degrees of satisfaction (or violation). An optimal solution is the complete assignment with an optimal satisfaction/violation degree. Therefore, choosing the operator used to perform the combination and an ordered satisfaction/violation scale is enough to define a specific framework.
Capturing these commonalities in a generic framework is desirable, since it allows us to design generic algorithms and properties instead of a myriad of apparently unrelated, but actually similar properties, theorems and algorithms. In Section~\ref{sec:soft} we show the \emph{semiring-based}  framework~\cite{jacm97,bistabook} that we will adopt in Section~\ref{sec:argsemi} in order to parametrize WAFs.

\subsection{Semiring-based Soft Constraints}\label{sec:soft}
A semiring~\cite{jacm97,bistabook} $S$ is a tuple $\langle
A,+,\times, \0, \1 \rangle$ where $A$ is a set with two special
elements $\0, \1 \in A$ (respectively the bottom and top elements of
$A$) and with two operations $+$ and $\times$ that satisfy certain
properties: $+$ is defined over (possibly infinite) sets of elements
of $A$ and is commutative, associative and idempotent; it is closed,
$\0$ is its unit element and $\1$  is its absorbing element;
$\times$ is closed, associative, commutative and distributes over
$+$, $\1$  is its unit element and $\0$ is its absorbing element. The
$+$ operation defines a partial order $\leq_S$ over $A$ such that $a
\leq_S b$ iff $a+b = b$; we say that $a \leq_S b$ if $b$ represents
a value \emph{better} \/than $a$. Moreover, $+$ and $\times$ are
monotone on $\leq_S$, $\0$ is its min and $\1$ its max, $\langle
A,\leq_S \rangle$ is a complete lattice and $+$ is its lub. Some practical instances of semirings are the \emph{Weighted semiring} $\langle \mathbb{R}^+ \cup \{\infty\}, \min, \hat{+}, \infty,
0\rangle$ ($\hat{+}$ is the arithmetic plus operation, to distinguish it from the generic semiring definition of $+$), the \emph{Fuzzy semiring} $\langle [0..1], \max, \min, 0,
1\rangle$, the \emph{Probabilistic semiring} $\langle [0..1], \max, \hat{\times}, 0,
1\rangle$ ($\hat{\times}$ is the arithmetic times operation, to distinguish it from the generic semiring definition of $\times$) and the \emph{Boolean semiring} $\langle \{true,
false\}, \vee, \wedge, false, true\rangle$, which can be used to model classical crisp CSPs.

Given  $S = \langle A,+,\times,\0,\1 \rangle$ and an
ordered set of variables $V$ over a finite domain $D$ (to simplify, we consider all the variables as defined on the same domain), a soft
constraint is a function which, given an assignment $\eta :
V\rightarrow D$ of the variables, returns a value of the semiring.
Using this notation  
$\C = \eta \rightarrow A$ is the set of all possible constraints
that can be built starting from $S$, $D$ and $V$.
Any function in $\C$ depends on the assignment of only a finite subset of $V$. For instance, a binary constraint $c_{x,y}$ over variables $x$ and $y$, is a function $c_{x,y}: V\rightarrow D\rightarrow A$, but it depends only on the assignment of variables $\{x,y\}\subseteq V$ (the  {\em scope}, of the constraint). 
Note that $c\eta[v:=d_1]$ means $c\eta'$ where $\eta'$ is $\eta$
modified with the assignment $v:=d_1$. 
Notice that $c\eta$ is the application of a constraint function $c:V
\rightarrow D \rightarrow A$ to a function $\eta:V\rightarrow D$;
what we obtain is a semiring value $c\eta=a$. Given the set $\C$, the combination
function $\otimes: \C\times\C \rightarrow \C$ is defined as
$(c_1\otimes c_2)\eta = c_1\eta\times
c_2\eta$~\cite{jacm97,bistabook}. The $\otimes$ builds a new
constraint  which associates with each tuple of domain values for
such variables a semiring element which is obtained by multiplying
the elements associated by the original constraints to the
appropriate sub-tuples. 
Given a constraint $c \in \C$ and a variable
$v \in V$, the {\em projection}~\cite{jacm97,bistabook} of $c$ over
$V \backslash \{v\}$, written $c\Downarrow_{(V \backslash\{v\})}$ is the
constraint $c'$ such that $c'\eta = \sum_{d \in D} c \eta [v:=d]$.
Informally, projecting means eliminating some variables from the scope.

A SCSP~\cite{bistabook} is defined as $P= \langle V, D, C, S \rangle$,
where $C$ is the set of constraints defined over variables in $V$ (each with domain $D$), and whose preference is determined by semiring $S$. The {\em best level of consistency}
notion is defined as $blevel(P) = Sol(P) \Downarrow_{\emptyset}$,
where $Sol(P)= \bigotimes C$~\cite{bistabook}. A problem $P$ is $\alpha$-consistent if
$blevel(P) = \alpha$~\cite{bistabook}. $P$ is instead simply
``consistent'' iff there exists $\alpha
>_S \0$ such that $P$ is $\alpha$-consistent~\cite{bistabook}. $P$ is inconsistent
if it is not consistent.

\begin{figure}[h]
\centering
\includegraphics[scale=0.6]{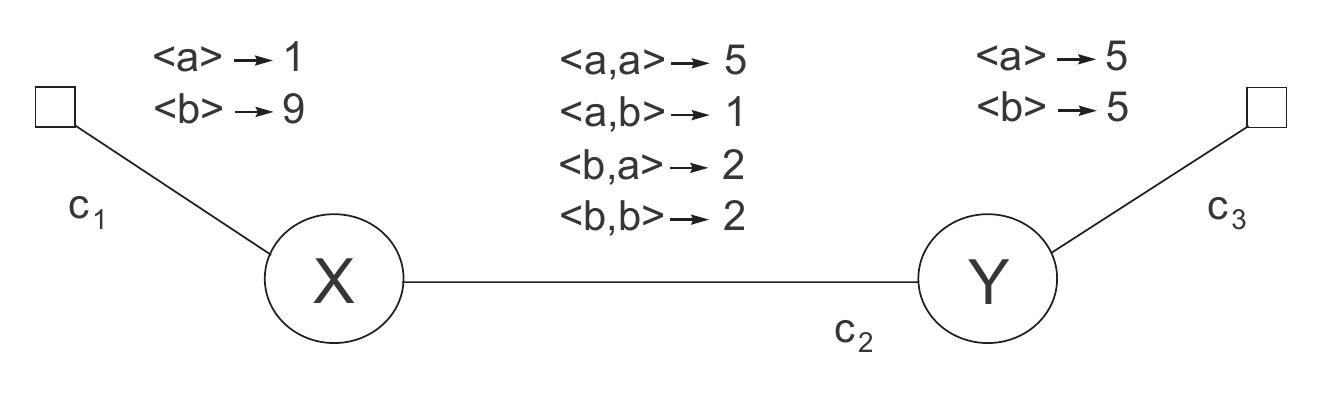} 
\caption{A SCSP based on a Weighted semiring.}
\label{figure:wexample}
\end{figure}

\begin{example}Figure~\ref{figure:wexample} shows a
weighted SCSP as a graph: the \emph{Weighted} semiring is used, i.e.
$\langle \mathbb{R}^+ \cup \{\infty\}, \min, \hat{+},$ $\infty,
0\rangle$ ($\hat{+}$ is the arithmetic plus operation). Variables
and constraints are represented respectively by nodes and arcs
(unary for $c_1$-$c_3$, and binary for $c_2$); $D= \{a, b\}$. The
solution of the CSP in Figure~\ref{figure:wexample} associates a
semiring element to every domain value of variables $X$ and $Y$ by
combining all the constraints together, i.e. $Sol(P)= \bigotimes C$.
For instance, for the tuple $\langle a, a\rangle$ (that is, $X = Y =
a$), we have to compute the sum of $1$ (which is the value assigned
to $X = a$ in constraint $c_1$), $5$ ($\langle X = a, Y = a \rangle$ in $c_2$) and $5$ ($Y = a$ in $c_3$): the value for this tuple is
$11$.  The \emph{blevel} is $7$, related to the solution $X = a$,
$Y = b$.
\end{example}

\section{Mapping AFs to CSPs}\label{mapping}

In this section we propose the mapping from AFs  to CSPs, which is at the hearth of ConArg. 
Given an $AF= \langle \mathcal{A}_{rgs}, R \rangle$, we
define a variable for each argument $a_i \in \mathcal{A}_{rgs}$
($V= \{a_1, a_2, \dots, a_n\}$) and each of these argument can
be taken or not, i.e., the domain of each variable is $D= \{1,
0\}$, $1$ if taken in the extension, $0$ if not taken.

 In the following explanation, notice
that $b$ attacks $a$ means that $b$ is a parent of $a$ in the
interaction graph, and $c$ attacks $b$ attacks $a$ means that $c$
is a grandparent of $a$. We need to define different sets of constraints:

\begin{enumerate}
\item \textbf{Conflict-free constraints.} Since we want to find the conflict-free sets, if $R(a_i, a_j)$ is in the graph we
need to prevent the solution to include both $a_i$ and $a_j$ in
the considered extension: $\neg  (a_i = 1\wedge a_j = 1)$.
The other possible assignment of the variables $(a= 0 \wedge b=1)$,
$(a=1 \wedge b= 0)$ and $(a= 0 \wedge b= 0)$ are permitted: in these cases we are choosing only one
argument between the two (or none of the two) and thus, we have no
conflict.
\item \textbf{Admissible constraints.} For the admissibility, we
need that if child argument $a_i$ has a parent  $a_p$, but $a_i$
has no grandparent  $a_g$ (parent of $a_p$), then we must avoid to take $a_i$ in
the extension because it is attacked and it cannot be defended by any
grandparent: this can be expressed with a unary constraint $a_i= 0$. 

Moreover, if $a_i$ has several grandparents $a_{g_1}, a_{g_2},\dots,
a_{g_k}$ and a parent $a_p$ (which is the child of $a_{g_1}, a_{g_2},\dots,
a_{g_k}$), we need to add a $k+1$-ary
constraint $\neg (a_i= 1 \wedge a_{g_1}= 0 \wedge \dots \wedge
a_{g_k}= 0)$. The explanation is that at least one grandparent
must be taken in the admissible set, in order to defend $a_i$ from
its parent $a_p$. Notice that, if an argument is not attacked
(i.e., it has no parents), it can be taken or not in the admissible
set.
\item \textbf{Complete constraints.} To compute a complete
extension $\mathcal{B}$, we impose that each  argument $a_i$ which
is defended by $\mathcal{B}$ is in $\mathcal{B}$, except those $a_i$ that, in such case, would be attacked by $\mathcal{B}$ itself~\cite{argcomputa}. This can be enforced by imposing that for each $a_i$ taken in the extension, also all its $k$ grandchildren $a_{s_1}, a_{s_2},\dots,
a_{s_k}$  (i.e., all the  arguments defended by $a_i$) whose parents  are not taken in the extension, must be  in $\mathcal{B}$. Formally, we enforce the assignments $(a_i= 1\wedge a_{s_1}= 1 \wedge \dots \wedge
a_{\mathit{s_k}}= 1)$ only for those $a_{\mathit{s_i}}$ for which it stands that $(a_{\mathit{p}_1}= 0 \wedge a_{\mathit{p}_2}= 0 \wedge \dots \wedge
a_{\mathit{p}_h}= 0)$, where $a_{\mathit{p}_1}, a_{\mathit{p}_2},\dots,
a_{\mathit{p}_h}$ are the $h$ parents of  $a_{s_i}$.
\item \textbf{Stable constraints.} If we have an argument $a_i$ with
$k$ parents $a_{p_1}, a_{p_2},$ $\dots, a_{p_k}$, we need to add
the constraint $\neg (a_i= 0 \wedge a_{p_1}= 0 \wedge
\dots \wedge a_{p_k}= 0)$. In words, if an argument is not taken in the
extension (i.e., $a_i= 0$), then it must be attacked by at least
one of the taken arguments: at least one parent of $a_i$ needs
to be taken in the  extension (i.e., $\exists j\in 1..k. \,a_{p_j} = 1$).

Moreover, if an argument $a_i$ has no parent in the graph, it has to be
included  in the stable extension; notice that $a_i$ cannot be attacked
by arguments inside the extension, since it has no parent. The
corresponding unary constraint is $\neg (a_i = 0)$.

\end{enumerate}

The
following proposition states the equivalence between solving an
$AF_S$ and its related CSP.

\begin{proposition}[Solution equivalence]\label{prop4}
Given an $AF= \langle \mathcal{A}_{rgs}, R\rangle$, the solutions of the
related $CSP$ (see Section~\ref{sec:softbg}) $P= \langle \mathcal{A}_{rgs}, \{0,1\}, C\rangle$ correspond to  all the
\begin{itemize}
\item conflict-free extensions by
using  $C = \{\textit{conflict-free}\}$ constraints,
\item admissible extensions by
using $C = \{\textit{conflict-free} \, \cup \, \textit{admissible}\}$ constraints,
\item complete extensions by
using $C = \{\textit{conflict-free} \, \cup \, \textit{admissible} \, \cup \, \textit{complete}\}$ constraints,
\item stable extensions by
using $C = \{\textit{conflict-free} \, \cup \, \textit{stable}\}$ constraints.
\end{itemize}
\end{proposition}

\paragraph{Grounded and preferred extensions} Concerning the other two classical semantics of Dung, i.e., the grounded and preferred ones, the solutions are obtained through two different steps: \emph{i)} first, all the complete (for the grounded case) and admissible (for the preferred case) extensions are obtained by solving the corresponding constraints given in Proposition~\ref{prop4}, \emph{ii)} then  these solutions are copied into a second CSP, where each variable has JaCoP type $\mathit{SetVar}$, that is defined as an ordered collection of integers. For instance, given $\mathcal{A}_{rgs} = \{a,b,c,d,e\}$, the admissible extension $\{a,c,d\}$ is translated into a $\mathit{SetVar}$ variable $\{1,0,1,1,0\}$. Then we add a constraint $\mathit{AinB}(X,Y)$  for each couple of these variables, which checks if variable $X$ is contained into variable $Y$; if this is true for at least one $Y$, then $X$ cannot be a preferred extension, otherwise, it corresponds to a preferred extension. Viceversa, if we first find all the complete extensions and then we impose a constraint $\mathit{AinB}(X,Y)$  for each couple of  variables, if $X$ is contained in each $Y$, this means that $X$ is a grounded extension.

\paragraph{Hard problems related to preferred extensions}
An interesting problem is determining whether a set of arguments $T$ is a preferred extension, which is a co-NP-complete~\cite{argcomputa} problem. In ConArg we explicitly offer to the user the opportunity to solve this problem as a CSP, which is made of less constraints than the one that searches for all the preferred extensions. In this particular case, \emph{i)} we still find all the admissible extensions, \emph{ii)} but after this we impose a $\mathit{AinB}(T,Y)$ for each admissible solution $Y$.
 
\paragraph{Semi-stable, stage and ideal extensions} The solution of these three extensions involves the computation of, respectively,  all the complete, conflict-free, and admissible/preferred extensions. For instance, to find semi-stable extensions, we need to add  conflict-free, admissible, and complete constraint classes to the problem, as defined in Proposition~\ref{prop4}. 

Furthermore, we need to add the constraints limiting an extension $\mathcal{E}$ according to its range, defined as $\mathcal{E} \cup \mathcal{E}^{+}$,  where $\mathcal{E}^{+}= \{a \in \mathcal{A}_{rgs} : \mathcal{E} \text{ attacks } a\}$ (see Section~\ref{sec:argurelated}). In order to find the range of an extension, we add $|\mathcal{A}_{rgs}|$ new \emph{Integer} variables, which are set to $1$ if the represented argument is attacked by at least one argument taken in the  complete extension. This is achieved by using the JaCoP conditional constraint $\mathit{IfThenElse}$, whose guard is represented by an $\mathit{Or}$ constraint (true if one of the parents is taken in the complete extension), and which sets the value of the new variables to $1$ or $0$ by using the $\mathit{XeqC}$ constraint (variable equals to constant value). Then, all the obtained solutions are translated into $\mathit{SetVar}$ variables, and maximality (w.r.t. set inclusion) is treated as for the preferred case, i.e., by solving a second CSP. 

The same procedure is used  for stage extensions as well, this time using admissible extensions as groundwork, instead of complete ones. Concerning the ideal semantics, we 

\begin{itemize} \item first find all the admissible extensions, and afterwards, by elaborating on these results, we find all the preferred extensions through the second step of the same CSP (preferred extensions are also admissible). \item Subsequently, we define a second CSP to check the precondition of the ideal semantics, that is if an admissible extension is subset of all the preferred extensions (see Def.~\ref{def6} in Section~\ref{sec:argurelated}). This second CSP receives the admissible and preferred extensions as input, obtained in the first CSP. In this second CSP, we impose that an admissible extension cannot be considered if it has an argument that is not taken in all the preferred extensions: in this way, we select only the admissible extensions that are subsets of the intersection of all the preferred extensions. To achieve this, we impose conflict-free and admissible constraint classes in order to find admissible solutions (see Proposition~\ref{prop4}), but we  also impose conditional $\mathit{IfThenElse}$ constraints to exclude admissible extensions that contain an argument which is not included in the intersection of all preferred extensions: $\mathit{And}$ constraints are used as guards, being ``false'' if an argument is not set to $1$ in all the preferred extensions. If this happens, a $\mathit{XeqC}$ forces the exclusion of that argument from the solution (i.e., it is se to $0$). 
\item Eventually, we deal with maximality (w.r.t. set inclusion) by translating the results of the second CSP into a third CSP, and applying the same solution adopted above for preferred/semi-stable/stage extensions. The solution of this third CSP corresponds to the ideal semantics.
\end{itemize}

\paragraph{Additional user-defined constraints} Notice that we can easily impose further requirements on the sets of arguments which are expected as extensions, like ``extensions must contain argument $a$ when they contain $b$'' or ``extensions must not contain one of $c$ or $d$ when they contain $a$ but do not contain $b$''~\cite{constrainedarg}. For example, with JaCoP it is straightforward to model this kind of side-requirements with conditional constraints as $\mathit{IfThen}(c_1,c_2)$, where constraint $c_2$ (e.g., extension contains argument $a$, that is $a=1$) must be satisfied if $c_1$ is satisfied (e.g., extension contains argument $b$, that is $b=1$). For the second example above, $c_1$ corresponds to $a=1 \wedge b=0$ and $c_2$ corresponds to $(c=1 \vee d=1) \wedge \neg(c=1 \wedge d=1)$.

To conclude, we remind that ConArg can solve all the problems presented in this section, among others. In Section~\ref{sec:argsemi}, we propose a  general parametrical framework where to express WAFs~\cite{ecai10}.

\section{Expressing Weighted AFs with Semirings}
\label{sec:argsemi} There have been a number of proposals for extending Dung
framework~\cite{dung} in order to allow for more sophisticated modeling and
analysis of conflicting information. A common theme among some
of these proposals is the observation that not all arguments are
equal, and that the relative strength of the arguments needs to be
taken into account somehow~\cite{inconsistency,vaf,main2,fuzzyattack,mogdil,probArg}.
WAFs extend Dung-style abstract argumentation
systems by adding numeric weights to every edge in the attack
graph, intuitively corresponding to the strength of the attack, or
equivalently, how reluctant we would be to disregard it.
In literature, we can find preferences directly associated with arguments~\cite{mogdil} or, more frequently, with
attacks~\cite{vaf,inconsistency,main2,fuzzyattack,probArg}. In this work we focus on weights 
associated with the attack relationships. 

\begin{figure}
\centering
\includegraphics[scale=0.6]{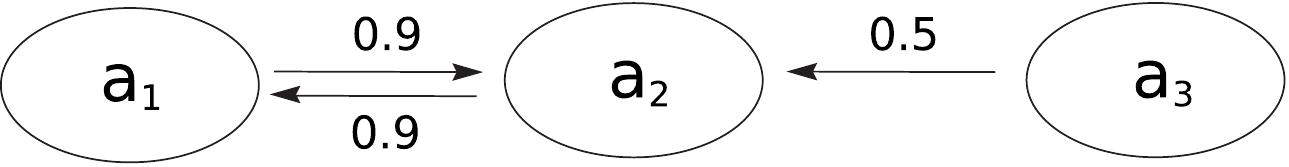} 
\caption{A fuzzy WAF, with fuzzy scores
modeling the attack strength among $a_1$ (\emph{``Today will be dry in London since
BBC forecast sunshine''}), $a_2$ (\emph{``Today will be
wet in London since CNN forecast rain''}), and $a_3$ (\emph{``BBC is more accurate that CNN''}).} \label{figure:strength}
\end{figure}

For example, in Figure~\ref{figure:strength} we represent a weighted interaction graph with three contradictory arguments about
weather forecasts announced by \emph{BBC} and \emph{CNN}:

\begin{description}
 \item[$\mathbf{a_1}$:] \small{Today will be dry in London since
BBC forecast sunshine.}  \item[$\mathbf{a_2}$:] \small{Today will be
wet in London since CNN forecast rain.} \item[$\mathbf{a_3}$:] \small{BBC is more accurate that CNN.}\end{description}

Therefore, we consider the following AF: $\mathcal{A}_{rgs}= \{a_1,a_2,a_3\}$, $a_1 \, R\, a_2$,   $a_2 \, R\, a_1$ and $a_3 \, R\, a_2$.
In Figure~\ref{figure:strength}, each of these three attack relationships is associated with
a fuzzy weight (in $[0, 1]$) representing the strength of the attack,  where $0$ represents the strongest possible attack, and $1$ the weakest one.

In the following we report how some works in literature can be  cast into the same parametrical semiring-based framework presented in Section~\ref{sec:soft}.

An argument can be seen as a chain of possible events that makes
the hypothesis true~\cite{probArg}. The credibility of a hypothesis can then be
measured by the total probability that it is supported by
arguments. To solve this problem we can use 
the \emph{Probabilistic} semiring $\langle
[0..1], max, \hat{\times}, 0, 1 \rangle$ (see Section~\ref{sec:soft}), where the arithmetic
multiplication (i.e., $\hat{\times}$) is used to compose the
probability values together (assuming that the probabilities being
composed are independent). In \cite{probArg} the authors associate probabilities with arguments and defeats. Then, they compute the likelihood of some set of arguments appearing within an arbitrary argument framework induced from the probabilistic framework. Weights can be also interpreted as subjective beliefs~\cite{main,main2}. For example, a weight of $w \in (0, 1]$
on the attack of argument $a_1$ on argument $a_2$ might be understood
as the belief that (a decision-maker considers) $a_2$ is false when $a_1$
is true. This belief could be modeled using probability~\cite{main,main2} as well.

The Fuzzy Argumentation approach presented in \cite{fuzzyarg} enriches the
expressive power of the classical argumentation model by allowing
to represent the relative strength of the attack relationships
between arguments, as well as the degree to which arguments are
accepted. In this case, the \emph{Fuzzy} semiring $\langle [0..1],
min, max, 0, 1 \rangle$ (see Section~\ref{sec:soft}) can be used, as for the example in Figure~\ref{figure:strength}.

In addition, the \emph{Weighted semiring} $\langle \mathbb{R}^+ \cup \{\infty\},
min, \hat{+}, \infty, 0 \rangle$ (where $\hat{+}$ is the arithmetic
plus) can model a generic ``cost'' for the attacks: for example,
the number of votes in support of the attack~\cite{main,main2}, which consequently need to be minimized. Other
possible interpretations of models that use the  \emph{Weighted semiring}  are provided in~\cite{main,main2}: for instance,  to rank the strengths of attacks in a relative way. 

With the \emph{Boolean semiring} $\langle \{true,
false\}, \vee, \wedge, false, true\rangle$ (see Section~\ref{sec:soft}), we can cast the classic
AFs originally defined by Dung~\cite{dung} in the same
semiring-based framework. Therefore, with a single parametrical semiring-based framework, we can capture the semantics of the different metrics used in literature by independent models. This leads to an unifying modeling framework, supported also by the solving techniques provided by (soft) Constraint Programming.



In the following of this section we rephrase all the classical definitions given in Section~\ref{sec:argurelated}, in order to parametrize them with the notions of semiring and weighted attacks. We call these new extensions as $\alpha$-extensions, because they tolerate a level $\alpha$ of  attack-strength  within the extension, while they they attack the arguments outside the coalition with more strength. This is the philosophy we used in designing these $\alpha$-extensions. 

The following definition rephrases the notion of WAF into \emph{semiring-based AF}, called $AF_S$:

\begin{definition}\label{def:waf}\emph{\textbf{(semiring-based AF)}}
A semiring-based Argumentation Framework ($AF_S$) is a quadruple
$\langle \mathcal{A}_{rgs}, R, W, S \rangle$, where $S$ is a semiring $
\langle A, +, \times, \0, \1 \rangle$, $\mathcal{A}_{rgs}$ is a set
of arguments, $R$ the attack binary relation  on
$\mathcal{A}_{rgs}$, and $W: \mathcal{A}_{rgs} \times \mathcal{A}_{rgs}
\longrightarrow A$ is a binary function  called the \emph{weight} function. Given $a,b \in \mathcal{A}_{rgs}$, $\forall (a, b)
\in R$, $W(a, b) = s$ means that $a$ attacks $b$ with a strength
level $s \in A$, the set of preference values of the semiring $S$.
\end{definition}

\begin{figure}
  \centering
    \includegraphics[scale=0.52]{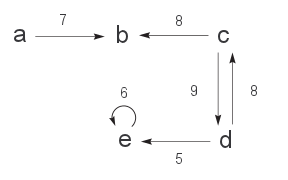}
    \caption{An example of a weighted interaction graph.}
    \label{fig:argnetex}
\end{figure}

In Figure~\ref{fig:argnetex} we provide an example of a weighted interaction graph describing the $AF_S$ defined by
$\mathcal{A}_{rgs}= \{a, b, c, d, e\}$, $R= \{(a, b), (c, b), (c, d), (d, c), (d, e), (e, e)\}$ with $W(a, b)= 7, W(c, b)= 8, W(c, d)= 9, W(d, c)= 8, W(d, e)= 5, W(e, e)= 6$ and $S= \langle \mathbb{R}^+ \cup \{\infty\}, \min, \hat{+}, \infty, 0\rangle$ (i.e., the \emph{Weighted} semiring). 

Therefore, each attack function is associated with a semiring
value that represents the ``strength'' of the attack between two
arguments. 
 We can consider the weights in Figure~\ref{fig:argnetex} as  votes supporting the associated attack. A semiring value equal to the top element of the semiring $\1$ (e.g., $0$ for the \emph{Weighted semiring}) represents  a no-attack relationship, not represented in Figure~\ref{fig:argnetex} to have a light notation. As a consequence of this, the bottom element of the semiring, i.e.,
$\0$ (e.g., $\infty$ for the \emph{Weighted semiring}), represents the strongest attack possible. 


In Def.~\ref{def:wset} we define the strength of attack for a set of arguments that attacks an argument or another set of arguments; in the following, we will use the product symbol $\displaystyle\prod^{S}$ in order to apply the $\times$ operator of the semiring $S$ on a sequence of semiring values:

\begin{definition}\label{def:wset}\emph{\textbf{(attacks for sets of arguments)}}
Given an $AF_S  = \langle \mathcal{A}_{rgs}, R, W, S \rangle$, a set of arguments $\mathcal{B}$ attacks an argument $a$ with a weight of $k$, that is $W(\mathcal{B},a)=k$, if $\displaystyle\prod_{b \in \mathcal{B}}^{S} W(b,a) = k$. A set of arguments $\mathcal{B}$ attacks a set of arguments $\mathcal{D}$ with a weight of $k$, that is $W(\mathcal{B},\mathcal{D})=k$, if $\displaystyle\prod_{b \in \mathcal{B}, d \in \mathcal{D}}^{S} W(b,d) = k$.
\end{definition}

In Def.~\ref{def:defw1} we redefine the notion of conflict-free set: conflicts can be now  part of the solution until a cost threshold $\alpha$ is met, and not worse: they are now called as $\alpha$-conflict-free solutions.

\begin{definition}\label{def:defw1}\emph{\textbf{($\alpha$-conflict-free extensions)}} Given an $AF_S  = \langle \mathcal{A}_{rgs}, R, W, S \rangle$, a subset of  arguments
$\mathcal{B} \subseteq \mathcal{A}_{rgs}$ is $\alpha$-conflict-free  iff $W(\mathcal{B}, \mathcal{B}) \geq_S \alpha$.\footnote{In case of a partially ordered semiring, the  $\geq_S$ is replaced by $\not<_S$. Similar considerations hold for the inequalities in the following of the text.}
\end{definition}

With respect to the $AF_S$ in Figure~\ref{fig:argnetex}, while the set $\{a, b, c\}$ is not conflict-free in the crisp version of the problem because it includes the attacks between $a$ and $b$ and between $c$ and $b$, $\{a, b, c\}$ is instead $15$-conflict-free  because $W(a,b) \hat{\times} W(c,b)= 15$.

We now define two propositions that derive from Definition~\ref{def:defw1} and the properties explained in Section~\ref{sec:soft}.

\begin{proposition}\label{prop:alpha}
If an extension is $\alpha_1$-conflict-free,  then the same extension is also $\alpha_2$-conflict-free if $\alpha_2 <_S \alpha_1$.
\end{proposition}

For instance,  $\{a, b, c\}$ is also a $17$-conflict-free because it is a $15$-conflict-free and $17 < 15$ in  the \emph{Weighted semiring}.


Definition~\ref{def:defastable} proposes  the Dung's stable extensions revisited in the semiring-based framework. 

\begin{definition}\label{def:defastable}\emph{\textbf{($\alpha$-stable extensions)}}
Given an $AF_S  = \langle \mathcal{A}_{rgs}, R, W, S \rangle$, an $\alpha$-conflict-free
set $\mathcal{B}$ is an $\alpha$-stable extension iff for each
argument $c \not\in \mathcal{B}$, $W(\mathcal{B},c) <_S \alpha$.
\end{definition}

For example, considering the problem in Figure~\ref{fig:argnetex} as unweighted (i.e., as a classical Dung AF), the set $\{a, d\}$ corresponds to the only stable extension. This set is also a $4$-stable extension, because it is $4$-conflict-free since it is $0$-conflict-free (see Proposition~\ref{prop:alpha}), and $b, c, e$ are attacked by an element in $\{a, d\}$ with a strength worse than $4$, that is $W(\{a,d\},b) = 7$,  $W(\{a,d\},c) = 8$, and $W(\{a,d\},e) = 5$. The extension $\{a,d,e\}$ is instead $11$-stable instead, since it is $11$-conflict-free and the other arguments $b$ and $c$ are attacked by at least one argument in $\{a,d,e\}$, i.e., $a R b$ and $d R c$. 

Like in Section~\ref{sec:softbg}, the other $\alpha$-extensions rely upon the concept of defense, in this case, \emph{weighted defense}:

\begin{definition}\label{defw2}\emph{\textbf{(weighted-defense)}} Given an $AF_S  = \langle \mathcal{A}_{rgs}, R, W, S \rangle$,
an argument $b \in \mathcal{A}_{rgs}$ is defended by a set $\mathcal{B} \subseteq
\mathcal{A}_{rgs}$ (or, $\mathcal{B}$ defends $b$) iff $\forall a \in \mathcal{A}_{rgs}$ such that $a R b$, we have that $W(a, b) >_S W(\mathcal{B}, a)$.
\end{definition}

The set $\{c\}$ in Figure~\ref{fig:argnetex} defends $c$ because $d R c$ and $W(d, c) >_S W(c, d)$, i.e., ($8 >_S 9$).\footnote{In the \emph{Weighted} semiring, $>_S$ is equivalent to $<$ over the Real numbers, in the \emph{Probabilistic} and \emph{Fuzzy} ones, $>_S$ corresponds to $>$ over the Real numbers in the interval $[0..1]$ (see Section~\ref{sec:soft}).} This definition reminds the notion of collective defeat presented in~\cite{abstractas,justdefeat}.
An $\alpha$-admissible set of arguments must be an
$\alpha$-conflict-free set that weighted-defends all its elements. Formally:

\begin{definition}\label{defw3}\emph{\textbf{($\alpha$-admissible extension)}}
Given an $AF_S  = \langle \mathcal{A}_{rgs}, R, W, S \rangle$, an $\alpha$-conflict-free set $\mathcal{B} \subseteq \mathcal{A}_{rgs}$ is
$\alpha$-admissible iff each argument in $\mathcal{B}$ is weighted-defended by
$\mathcal{B}$.
\end{definition}

Not considering weights in Figure~\ref{fig:argnetex}, the admissible sets are: $\{a\}, \{c\}, \{d\}, \{a, c\},$ $\{a,$ $ d\}$.  The  $\1$-admissible extensions  are $\{a\}$, $\{c\}$, and $\{a, c\}$ instead: $\{a\}$ because is not attacked by any other argument, $\{c\}$ and $\{a,c\}$ because $c$ is able to weighted-defend itself from the attack performed by $d$, i.e., $W(d, c) >_S W(c, d)$.  As a further example, $\{a, b, c\}$ is $15$-admissible because it is $15$-conflict-free, and $c$ weighted-defends himself from $d$, as explained before. All the $15$-admissible extensions are $\emptyset$, $\{c\}$, $\{c,e\}$, $\{a\}$, $\{a,c\}$, $\{a,c,e\}$, and $\{a,b,c\}$.

Besides the $\alpha$-stable semantics, three semantics refining
$\alpha$-admissibility can be introduced:

\begin{definition}\label{defw4}\emph{\textbf{($\alpha$-preferred, $\alpha$-complete and $\alpha$-grounded extensions)}}
An $\alpha$-preferred extension is a maximal (w.r.t. set inclusion)
$\alpha$-admissible subset of $\mathcal{A}_{rgs}$. An $\alpha$-admissible
$\mathcal{B} \subseteq \mathcal{A}_{rgs}$ is an $\alpha$-complete extension
iff each argument which is weighted-defended by $\mathcal{B}$ is in
$\mathcal{B}$. The least (w.r.t. set inclusion) $\alpha$-complete extension
is the $\alpha$-grounded extension.
\end{definition}

Note that now, if we can disregard consistency at will, we can always take the whole $\mathcal{A}_{rgs}$ set as
an admissible and then preferred extension: $\{a, b, c, d, e\}$ in Figure~\ref{fig:argnetex} the $43$-admissible extension of course maximal, i.e., it is also preferred.

In Def.~\ref{defw5} we redefine also the semi-stable semantics as proposed in~\cite{semistable}. According to~\cite{semistable}, given $a \in \mathcal{A}_{rgs}$ and $\mathcal{B} \subseteq \mathcal{A}_{rgs}$ we define $a_\alpha^+$ as $\{c \,|\, W(a,c)<_S \alpha\}$ and $\mathcal{B}_\alpha^+$ as $\{c \,|\, W(\mathcal{B}, c) <_S \alpha\}$.

\begin{definition}\label{defw5}\emph{\textbf{($\alpha$-semi-stable extension)}} Given
$AF_S= \langle \mathcal{A}_{rgs},R,$ $ W, S \rangle$ and $\mathcal{B} \subseteq \mathcal{A}_{rgs}$, $\mathcal{B}$ is
called an $\alpha$-semi-stable extension iff  $\mathcal{B}$  is $\alpha$-complete and $\mathcal{B} \cup \mathcal{B}_\alpha^+$, called the $\alpha$-range of $\mathcal{B}$,  is maximal w.r.t. set inclusion.
\end{definition}

Some classical~\cite{semistable,dung}  properties still hold among the new $\alpha$-extensions:

\begin{theorem}\label{th2}
Given
$AF_S= \langle \mathcal{A}_{rgs},R,$ $ W, S \rangle$, with a semiring $S = \langle A, +, \times, \0, \1 \rangle$, and an $\alpha \in A$, then
\begin{enumerate}
\item every $\alpha$-complete is also $\alpha$-admissible.
\item every $\alpha$-preferred extension is also $\alpha$-complete.
\item an $\alpha$-grounded extension is contained in every $\alpha$-preferred one.
\item every $\alpha$-stable extension is also  $\alpha$-semi-stable.
\item every $\alpha$-semi-stable extension is also $\alpha$-preferred.
\end{enumerate}
\end{theorem}

\emph{Proof.}  \emph{1)} is trivially proved by definition (see Def.~\ref{defw3} and Def.~\ref{defw4}). For point \emph{2)}, if $\mathcal{B}$ is the maximal set such that each argument in $\mathcal{B}$ is weighted-defended by
$\mathcal{B}$, then each argument which is weighted-defended by $\mathcal{B}$ is in $\mathcal{B}$ (i.e., $\mathcal{B}$ is $\alpha$-complete). \emph{3)}  derives from \emph{1)} and from the definition of  $\alpha$-grounded extension, which is the minimal (w.r.t. set inclusion) $\alpha$-complete extension. Concerning  \emph{4)}, by definition an $\alpha$-stable extension maximizes the $\alpha$-range (see Def.~\ref{defw5}), and, therefore, it is also $\alpha$-semi-stable. To prove \emph{5)},  let $\mathcal{B}$ be an $\alpha$-semi-stable extension. Suppose $\mathcal{B}$ is not an $\alpha$-preferred
extension, then there exists a set $\mathcal{B}' \supsetneq \mathcal{B}$ such that $\mathcal{B}'$ is
an $\alpha$-complete extension. It follows that $\mathcal{B}'^+ \supsetneq \mathcal{B}^+$. Therefore,
$(\mathcal{B}' \cup \mathcal{B}'^+) \supsetneq (\mathcal{B} \cup \mathcal{B}^+)$. But then  $\mathcal{B}$ would not be an $\alpha$-semi-stable
extension, since $\mathcal{B} \cup \mathcal{B}^+$ would not be maximal, and this leads to a contradiction. \qed

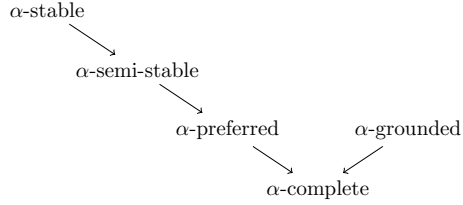
\begin{figure}
  \centering
  \centering\scalebox{0.8}{
\begin{tikzpicture}
    \tikzstyle{all nodes}=[inner sep=0pt]
    \draw 
              node(A)at(1.5,1){$\alpha$-stable}
              node(B)at(3,0){$\alpha$-semi-stable}
              node(C)at(4.5,-1){$\alpha$-preferred}
              node(D)at(6,-2){$\alpha$-complete}
              node(E)at(7.5,-1){$\alpha$-grounded};

  \foreach\xa/\xb in{A/B,B/C,C/D,E/D}
      {\draw[->](\xa)--(\xb);}
\end{tikzpicture}}
    \caption{Inclusion-relationship hierarchy between $\alpha$-extensions.}
    \label{fig:inclusions}
\end{figure}

Theorem~\ref{th2} leads to  Corollary~\ref{cor1}, which states that the classical inclusion relationships between extensions~\cite{semistable,dung} still hold also in our weighted framework. This is also represented in Figure~\ref{fig:inclusions}.

\begin{corollary}\label{cor1}
The following general inclusion relationships hold between $\alpha$-extensions: $\alpha$-stable $\subseteq \alpha$-semi-stable $\subseteq$ $\alpha$-preferred   $\subseteq$ $\alpha$-complete, and  $\alpha$-grounded $\subseteq
$ $\alpha$-complete.
\end{corollary}

Theorem~\ref{th3} relates the new  $\alpha$-extensions to their counterpart in the classical Dung's framework~\cite{dung}. 

\begin{theorem}\label{th3}
Given a classical $\mathit{AF} = \langle
\mathcal{A}_{rgs}, R \rangle$ as defined in Def.~\ref {def1}, and 
any possible related $\alpha$-version of it $AF_S= \langle \mathcal{A}_{rgs},R,$ $ W, S \rangle$, then
\begin{enumerate}
\item $\1$-conflict-free extensions in $AF_S$  correspond to conflict-free ones in $AF$.
\item $\1$-admissible extensions in $AF_S$  are a subset of admissible ones in $AF$.
\item $\1$-complete extensions in $AF_S$  are a subset of complete ones in $AF$.
\item $\1$-semi-stable extensions in $AF_S$  are equivalent to semi-stable ones in $AF$.
\item $\1$-stable extensions in $AF_S$  correspond to stable ones in $AF$.
\item $\1$-grounded extensions in $AF_S$  are a subset of grounded ones in $AF$.
\item $\1$-preferred extensions in $AF_S$  are a subset of preferred ones in $AF$.
\end{enumerate}
\end{theorem}

\emph{Proof.} Concerning \emph{1)}, a semiring value equal to the top element of the semiring (i.e., $\1$)  represents a no-attack relationship, so $\alpha$-conflict-free extensions do not include any attack among their arguments (i.e., they are conflict-free~\cite{dung}). \emph{2)} and \emph{3)} hold because the notion of weighted-defense (see Def.~\ref{defw2}) implies the classical notion of defense (see Def.~\ref{def2}). \emph{4)} and \emph{5)} hold because, if the taken arguments attack all, or maximize, the arguments outside with a strength greater that $\1$ (i.e., they are $\alpha$-stable and $\alpha$-semi-stable), it means that they respectively are stable and semi-stable according to the not-weighted semantics~\cite{semistable,dung} hold. \emph{6)} and \emph{7)} can be respectively proved after \emph{3)} and \emph{2)}. \qed

At last, note that the cartesian product of two semirings is
still a semiring~\cite{jacm97,bistabook}, and this can be
fruitfully used to describe multi-criteria constraint optimisation problems.

\section{Mapping Weighted $AF_S$ to a SCSP.}\label{sec:mapping2}

In this section we propose a mapping from semiring-based AF, that is the $AF_S$ presented in Section~\ref{sec:argsemi}, to semiring-based SCSPs (see Section~\ref{sec:soft}), as we do in Section~\ref{mapping} for not-weighted AF~\cite{dung}: in this way, we can find all the $\alpha$-extensions described in Section~\ref{sec:argsemi} as a solution of the corresponding SCSP.

Given an $AF_S= \langle \mathcal{A}_{rgs}, R, W, S \rangle$ over a semiring $S = \langle A, +, \times, \0, \1\rangle$ (see Section~\ref{sec:argsemi}), we
define a variable for each argument $a_i \in \mathcal{A}_{rgs}$,
that is  $V= \{a_1, a_2, \dots, a_n\}$ and each of these argument can
be taken or not as an element of one of the $\alpha$-extensions, i.e., the domain of each variable is $D= \{1,
0\}$: $1$ when the element belongs to the $\alpha$-extension, $0$ otherwise. Parent and child relationships among arguments are used in the following formulation as in Section~\ref{mapping}, that is considering the corresponding weighted interaction-graph. To compute the different $\alpha$-extensions we need to define distinct sets of constraints:

\begin{enumerate}

\item \textbf{$\alpha$-conflict-free constraints.} Since we want to find $\alpha$-conflict-free extensions, if $W(a_i, a_j)= s <_S \1$ ($s \in A$)  we
need assign a $s$ ``cost'' to the solution that includes both $a_i$ and $a_j$ in
the considered $\alpha$-conflict free extension: $c_{a_i,a_j} (a_i = 1, a_j = 1) = s$.
For the other possible variable assignments (i.e., $(a= 0, b=1)
(a=1, b= 0)$ and $(a= 0, b= 0)$), $c_{a_i,a_j}= \1$, since no conflict is introduced in the extension.

\item \textbf{$\alpha$-admissible constraints.} For the admissibility, we
need that, if a child argument $a_i$ has a parent  $a_p$, but $a_i$
has no grandparent  $a_g$, then we must avoid to take $a_i$ in
the extension because it is attacked and cannot be defended by any
grandparent: this can be expressed with a binary constraint, $c_{a_p,a_i}(a_p= 0,
a_i= 1)= \0$, which is equal to $\1$ for the other assignments of $a_p$ and $a_i$. Note that, differently from crisp admissible constraints in Section~\ref{mapping}, here the assignment
$c_{a_p,a_i}(a_f= 1, a_i= 1)$ is allowed (it has a preference value of $\1$) because we tolerate attacks inside an $\alpha$-extension.

Moreover,  we need to add a $k+1$-ary
constraint  $c_{a_i,a_{g_1},\dots, a_{g_k}}(a_i=1, a_{g_1}=X_1, \ldots, a_{g_k}=X_k)$ among an argument $a_i$ and its $k$ grandparents $a_{g_i}$,
where each $X_i \in D = \{0,1\}$, that is each grandparent can be taken in the $\alpha$-admissible set or not ($0/1$ respectively). The preference for this constraint is equal to $\0$ if
$$\prod_{g_i \;  \emph{for}  \; i= 1..k, \; \emph{s.t.} \, X_i =1}  W(a_{g_i},a_{p}) >_S  W(a_{p},a_{i})$$
, or equal to $\1$ otherwise (i.e., if $\leq_S$).
In words, the constraint has a preference value of $\0$ if the composition of the attack-weights of the taken grandparents towards a parent $a_p$ of $a_i$ is weaker than the attack of $a_p$ towards $a_i$. This because, as defined in Definition~\ref{defw3}, this composition has to be stronger or equal, according to the preference-ordering of the adopted semiring (concept of weighted-defense, see Definition~\ref{defw2}).
\item \textbf{$\alpha$-complete constraints.} To compute a complete
extension $\mathcal{B}$, we impose that each  argument $a_i$ that
is defended by $\mathcal{B}$ is in $\mathcal{B}$, except those $a_i$ that, in such case, would be attacked by $\mathcal{B}$ itself~\cite{argcomputa}. This can be enforced by imposing that for each $a_i$ taken in the extension, also all its $k$ grandchildren $a_{s_1}, a_{s_2},\dots,
a_{s_k}$  (i.e., all the  arguments defended by $a_i$) whose parents  are not taken in the extension, must be  in $\mathcal{B}$. Formally, $c_{a_i,a_{s_1},\dots, a_{s_k}}(a_i= 1, a_{s_1}= 1 , \dots,
a_{s_k}= 1)= \1$ only for those $a_{s_i}$ for which it stands that $(a_{p_1}= 0, a_{p_2}= 0, \dots,
a_{p_h}= 0)$, where $a_{p_1}, a_{p_2},\dots,
a_{p_h}$ are the $h$ parents of  $a_{s_i}$; otherwise, $c_{a_i,a_{s_1},\dots, a_{s_k}}= \0$. Notice that the condition of weighted-defense for $\alpha$-complete extensions is granted by imposing also $\alpha$-admissible constraints in the problem (see Proposition~\ref{th}).
\item \textbf{$\alpha$-stable constraints.} If we have a child node $a_i$ with
multiple parents $a_{f1}, a_{f2},$ $\dots, a_{fk}$, we need to add
the constraint $c_{a_i,a_{f1},\dots, a_{fk}}(a_i= 0, a_{f1}= 0,
\dots, a_{fk}= 0)= \0$. In words, if a node is not taken in the
extension (i.e., $a_i= 0$), then it must be attacked by at least
one of the taken nodes, that is at least a parent of $a_i$ needs
to be taken in the stable extension (that is, $a_{fj} = 1)$.

Moreover, if a node $a_i$ has no parent in the graph, it has to be
included  in the stable extension (notice $a_i$ cannot be attacked
by nodes inside the extension, since he has no parent). The
corresponding unary constraint is $c_{a_i} (a_i = 0) = \0$.

\end{enumerate}

Proposition~\ref{th} shows how to find all the $\alpha$-extensions presented in Section~\ref{sec:argsemi}, by using the proper classes of  constraints to build the intended \emph{SCSP}:

\begin{proposition}\label{th}\emph{\textbf{(Equivalence for $\alpha$-extensions)}}
Given a semiring-based Argumentation Framework $AF_S= \langle \mathcal{A}_{rgs}, R, W, S \rangle$ and the related $SCSP$ (see Section~\ref{sec:soft}) $P= \langle \mathcal{A}_{rgs}, \{0,1\}, C, S\rangle$, the $\alpha$-consistent solutions of $P$ (see Section~\ref{sec:soft}) corresponds to all the
\begin{itemize}
\item $\alpha$-conflict-free extensions by
using  $C = \{\textit{$\alpha$-conflict-free}\}$ constraints,
\item $\alpha$-admissible extensions by
using $C = \{\textit{$\alpha$-conflict-free} \, \cup \, \textit{$\alpha$-admissible}\}$ constraints,
\item $\alpha$-complete extensions by
using $C = \{\textit{$\alpha$-conflict-free} \, \cup  \textit{$\alpha$-admissible}  \cup \, \textit{$\alpha$-complete}\}$ constraints,
\item $\alpha$-stable extensions by
using $C = \{\textit{$\alpha$-conflict-free} \, \cup \, \textit{$\alpha$-stable}\}$  constraints.
\end{itemize}
\end{proposition}

These constraints have been implemented in JaCoP similarly as described for their not-weighted versions in Section~\ref{mapping}. To deal with costs, differently from Section~\ref{mapping}, we introduce a new $\mathit{IntVar}$ variable to represent the cost of an attack. In Figure~\ref{fig:jacopcode} we present the JaCoP code we use to find $\alpha$-conflict-free extensions. The first  $\mathit{IfThenElse}$ constraint is used to specify the cost of an argument $a_i$ attacking $a_j$. This cost, which is saved in the new $\mathit{IntVar}$ variable $\mathit{costArray}[k]$, is equal to  the cost of the attack between $a_i$ and $a_j$ (i.e., $\mathit{attackCost}[i,j]$) if both $a_i$ and $a_j$ are taken in the extension (i.e, the are both equal to $1$). Otherwise (\emph{else} branch) it is equal to the top preference of the semiring: in this case of \emph{Weighted} semiring, it is equal to $0$. The same constraint is repeated for each pair of attacking arguments in the weighted interaction graph. At last, the sum (i.e., \emph{Sum} constraint in Figure~\ref{fig:jacopcode}) of all these costs is computed in the variable $\mathit{totalCost}$, which is imposed to be less or equal to $\alpha$ (i.e., \emph{XlteqC} constraint Figure~\ref{fig:jacopcode}).

The conditions on the attack costs for the other classes of constraints, that is $\alpha$-admissible, $\alpha$-complete, and $\alpha$-stable ones, are managed in the same way: we add variables to represent the costs, and we constrain the value of the their sum to be less/equal than a threshold. As regards $\alpha$-grounded and $\alpha$-preferred extensions, minimality and maximality with respect set inclusion are solved as explained in Section~\ref{sec:mapping2} for their not-weighted version.

\begin{figure}
\begin{center}\footnotesize
{\tt /* $\dots$ for each argument $a_i$ attacking an argument $a_j$ $\dots$ */}\\
{\tt store.impose(new IfThenElse(new And(new XeqC(a[i], 1), new XeqC(a[j], 1)), new XeqC(costArray[k], attackCost[i,j]), new XeqC(costArray[k], 0])));}\\
{\tt /*  To impose that the total cost of the attacks is below threshold $\alpha$ */}\\
{\tt  store.impose(new Sum(costArray, totalCost));}\\
{\tt  store.impose(new XlteqC(totalCost, alpha));}
\caption{An example of JaCoP code to find $\alpha$-conflict-free extensions.} \label{fig:jacopcode}
\end{center}
\end{figure}

In ConArg we have implemented two different semirings from Section~\ref{sec:soft}, that is the \emph{Weighted} semiring  $\langle \mathbb{R}^+ \cup \{\infty\}, \min, \hat{+}, \infty, 0\rangle$ and the \emph{Fuzzy} semiring $\langle [0..1], min, max, 0, 1 \rangle$. Therefore, it is possible find all the $\alpha$-extensions presented in this section according to these two different system of preferences. To conclude, we remind that ConArg can find all the $\alpha$-extensions  in this section.

\subsection{Weighted Grounded Extensions~\cite{main2}}\label{sec:bgrounded}
ConArg is also able to solve hard problems related to the WAF formalism presented in \cite{main,main2}. More precisely, we can find all the $\beta$-grounded extensions (see Section~\ref{sec:bgweightedAF}), and we can also give a solution to all the problems described in Proposition~\ref{prop1}, Proposition~\ref{prop2} and Proposition~\ref{prop3} (see Section~\ref{sec:bgweightedAF}). We have decided to include also these problems because they are NP-complete (i.e., Proposition~\ref{prop1}) or co-NP-complete (i.e., Proposition~\ref{prop2} and Proposition~\ref{prop3}), and we can consequently take advantage of Constraint Programming~\cite{bookrossi} to tackle their inherent complexity.

In~\cite{main,main2}, $\beta$-grounded extensions are computed as Dung's classical grounded-extensions~\cite{dung}, but only after having removed from the WAF all the attacks whose strengths sum up to an inconsistency budget defined by a threshold $\beta$. Therefore, from an original WAF and a given $\beta$ we can obtain several derived WAFs, on which classical grounded extensions are computed.

As a result, in our implementation of ConArg we find $\beta$-grounded extensions exactly as described in Section~\ref{mapping} for classical grounded extensions, that is with complete, admissible and complete constraints, and then by checking the minimality with respect set inclusion. Threshold $\beta$ is given as input from a user. To solve the problem explained in Proposition~\ref{prop1}, we impose the value of the input argument $a$ as equal to $1$ (i.e., $a$ must be present in the extension), by using JaCoP constraint \emph{XeqC}.  Then we can state that the problem has a solution as soon as we find a $\beta$-grounded extension (containing $a$), or no solution otherwise. To solve the problem described in Proposition~\ref{prop2} , we proceed in the same way as for Proposition~\ref{prop1}, but we require that $a$ is contained in each $\beta$-grounded extension; in this case, the problem is positively solved. The problem in Proposition~\ref{prop3} is slightly more complex: to solve it, first we check that the input set $L$ correspond to a $\beta$-grounded extension. Since we can solve minimality between a set and a set of sets (we do the same to find grounded extension, for example), afterwards we check if $L$ w.r.t. all the other $\beta$-grounded extensions.

\section{The  Tool}\label{ConArg}
In this section we briefly present the visual interface and all the options of ConArg,\footnote{Downloadable at \url{https://sites.google.com/site/santinifrancesco/tools/ConArg.zip}} our visual tool that  generates interaction graphs and finds Dung's extensions~\cite{dung} over it (see Section~\ref{sec:argurelated}) by using Constraint Programming~\cite{bookrossi}. ConArg has been entirely programmed in the Java language using the \emph{NetBeans} development environment,\footnote{\url{http://netbeans.org/}}. ConArg can be downloaded as an archive file containing the \emph{.jar} file of the project, and a directory with all the \emph{.jar} files of the used third-party libraries. 

To program and solve constraints we adopted the \emph{Java Constraint Programming} library (JaCoP), which is a Java library that provides the user with \emph{Finite
Domain Constraint
Programming} paradigm~\cite{bookrossi}. JaCoP provides different type of constraints:
for example, the most commonly used primitive constraints, such as arithmetical
constraints, equalities and inequalities, logical, reified and
conditional constraints, combinatorial (global) constraints. It provides a significant number of (global) 
constraints to facilitate an efficient modeling. Finally, JaCoP defines also decomposable constraints, i.e., constraints that are defined using other constraints and possibly auxiliary variables. It also provides a modular design of search to help the user on specific characteristics of the problem being addressed.

\begin{figure}
  \centering
    \includegraphics[scale=0.47]{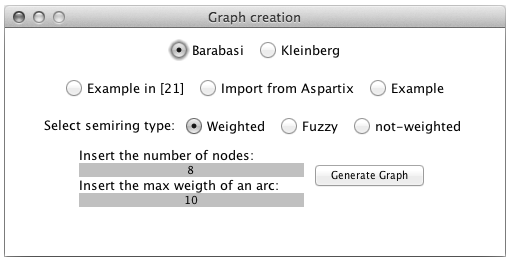}
    \caption{The window of ConArg where it is possible to generate the desired interaction graph.}
    \label{fig:genGraph}
\end{figure}

The fist window of the graphical interface of ConArg can be used to choose the interaction graph we want to adopt to solve our argumentation-related problems; it is depicted in Figure~\ref{fig:genGraph}. To  generate and work with these graphs  we use the \emph{Java
Universal Network/Graph Framework} (\emph{JUNG})~\cite{JUNG}, a Java
software library for the modeling, generation, analysis and
visualization of graphs. With JUNG we are capable to generate directed graph, where nodes are 
considered as arguments, and edges as directed attacks. 

It is possible to generate five different kinds of interaction graphs: from left to right in Figure~\ref{fig:genGraph} (top to bottom), it is possible to select:

\begin{enumerate}[i)]
\item a random \emph{Barabasi} network with small-world properties~\cite{Barabasi}. In this case (and in \emph{ii)}) it is also possible to select the desired number of arguments/nodes in the generated network. The JUNG library~\cite{JUNG} implements a simple evolving scale-free random graph generator. At each time step, a new vertex is created and  connected to existing vertices according to the principle of ``preferential attachment''~\cite{Barabasi}, whereby vertices with higher degree have a higher probability of being selected for attachment. At a given time-step, the probability $p$ of creating an edge between an existing vertex $v$ and the newly added vertex is $p = (\mathit{degree}(v) + 1) / (|E| + |V|)$. $|E|$ and $|V|$ are, respectively, the number of edges and vertices currently in the network.
\item a random \emph{Kleinberg} network with small-world properties~\cite{kleinberg}. Kleinberg adds a number of directed long-range random links to an $n\times n$ lattice network (vertices as nodes of a grid, undirected edges between any two adjacent nodes). Links have a non-uniform distribution that favors arcs to close nodes over more distant ones. In the implementation provided by JUNG~\cite{JUNG}, each node $u$ has four local connections, one to each of its neighbors, and in addition one or more long range connections to some node $v$, where $v$ is chosen randomly according to probability proportional to $d^\theta$ where $d$ is the lattice distance between $u$ and $v$ and $\theta$ is the clustering exponent, which can be specified by a user in the window in Figure~\ref{fig:genGraph}. Note that the number of  nodes, which can be selected in Figure~\ref{fig:genGraph}, corresponds to $n$, leading to  a total of $n \times n$ nodes in the final generated graph.
\item the case-study interaction graph presented in~\cite{main,main2}, in order to let a user compare the solutions of ConArg with the same solutions given in \cite{main,main2}.
\item a textual description of the network, saved as a file with the \emph{.dl} extension. In this way it is possible to import a user's own network in ConArg. We decided to use the \emph{.dl} extension because, in this way, we are capable to import examples generated and used in ASPARTIX~\cite{aspartix}. This textual format is really easy to use, since, in its basic form, it only consists in a list of node and attack declarations: for example, the AF where $\mathcal{A}_{rgs} = \{a,b,c\}$ and $a R b$, $b R c$, consists in the file $\mathtt{arg(0).\,  arg(1).\, arg(2). \, att(0,1). \, att(1,2).\;}$.
\item the interaction graph represented in this paper in Figure~\ref{fig:argnetex}, in order to check the correctness of the examples reported in Section~\ref{sec:argsemi}.
\end{enumerate}

All the generated graphs can be then also exported to the same \emph{.dl} format used by ASPARTIX~\cite{aspartix}, but only in their not-weighted form. Therefore, it is possible to test ASPARTIX over the random graphs generated in case \emph{i)} and \emph{ii)}. Since these two kinds of graph are randomly generated, successive generations with the same exact parameter result in different output networks.

In the same window (see Figure~\ref{fig:genGraph}) it is possible to select the weights we can assign to attacks as well. Consequently, it is possible to mix previous options \emph{ i)-v)} with following options \emph{a)-c)}. From left to right in Figure~\ref{fig:genGraph} we can:

\begin{enumerate}[a)]
\item  randomly assign to arcs/attacks a weight in the interval $[1\dots \mathit{max}]$, where $\mathit{max}$ is selected by a user before the generation (see Figure~\ref{fig:genGraph}). These weights are then interpreted in the \emph{Weighted} semiring $\langle \mathbb{R}^+ \cup \{\infty\}, \min, \hat{+}, \infty, 0\rangle$ (see Section~\ref{sec:soft}).
\item  randomly assign to attacks a weight in the interval $[0\dots 1]$. These weights are then interpreted in the \emph{Fuzzy} semiring $\langle [0..1], min, max, 0, 1 \rangle$ (see Section~\ref{sec:soft}).

\item  generate no weight for the attack, in order to model classical Dung's AF~\cite{dung}.
\end{enumerate}

\begin{figure}
  \centering
    \includegraphics[scale=.43]{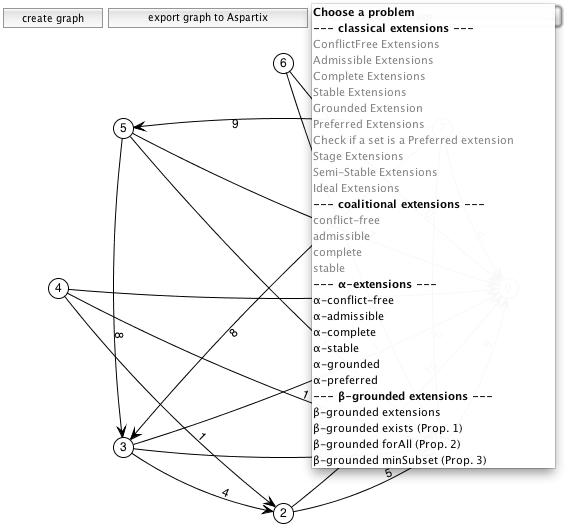}
    \caption{The drop-down list shows all the possible problems that can be solved in ConArg.}
    \label{fig:conargextensions}
\end{figure}

After the generation of the interaction graph, which becomes visible in the ConArg window together with the weights on the arcs (if required during the generation), it is possible to select the desired problem we want to solve.  This is illustrated in the drop-down list visualised in Figure~\ref{fig:conargextensions}. The problems are grouped by related topic, i.e., classical extensions (see Section~\ref{mapping}), coalitional extensions (described in~\cite{sac11}, but out of the scope of this paper), $\alpha$-extensions (see Section~\ref{sec:mapping2}), and problems related to $\beta$-grounded extensions (see Section~\ref{sec:bgrounded}). If the generated graph is weighted, then it is only possible to solve problems related to $\alpha$-extensions and $\beta$-grounded extensions, while if the graph is not weighted, it is only possible to solve classical problems~\cite{dung}.

By clicking on a specific problem is then possible to be asked for additional information, as for example an $\alpha$ threshold for $\alpha$-extensions (see Section~\ref{sec:argsemi}), or a $\beta$ threshold for $\beta$-grounded extensions~\cite{main,main2}. After the solutions are computed, a user can graphically browse all of them, where arguments/nodes taken in the solution (i.e., in the corresponding extension) are filled with color gray to be distinguished from arguments outside the extension.

\begin{figure}
  \centering
    \includegraphics[scale=.43]{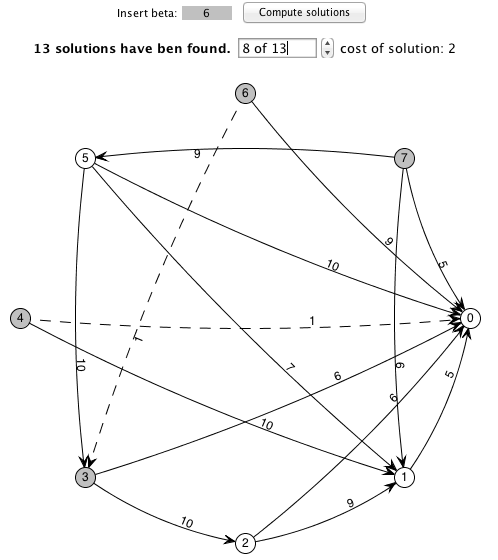}
    \caption{The drop-down list shows all the possible problems that can be solved in ConArg.}
    \label{fig:betaexample}
\end{figure}

 In Figure~\ref{fig:betaexample} we show the eighth  solution (out of thirteen)  after having asked to find all the $\beta$-grounded extensions with $\beta$ equal to $6$. The arguments/nodes taken in the extension correspond to argument id-numbers $3$, $4$, $6$ and $7$. The considered graph has been obtained by removing arcs corresponding to the attacks between $6$ and $3$, and $4$ and $0$, represented by the dotted lines in Figure~\ref{fig:betaexample}. This because in \cite{main,main2} it is possible to tolerate an inconsistency in the solution up to threshold $\beta$. The tolerated inconsistency corresponds to the sum of the weights on the removed arcs, which, in the case of  Figure~\ref{fig:betaexample}, is equal  to $2$ (as reported also in the figure).

\paragraph{Argumentation and social networks} In order to test ConArg over sensibly wide interaction-graphs (see Section~\ref{sec:tests}), our attention has turned to random networks with small-world properties, as \emph{Barabasi}~\cite{Barabasi} and \emph{Kleinberg}~\cite{kleinberg} networks.

The reason is that social networks usually show a structure typical of small-world graphs~\cite{discussion}. A practical example can be the study of discussion fora or 
discussion groups, where the users post their arguments that may attack other users' arguments. Everyday examples are online social platforms, such as 
\emph{Facebook}\footnote{\url{http://www.facebook.com}}, e-commerce sites, such as \emph{Amazon}\footnote{\url{http://www.amazon.com}},
and technical fora, such as \emph{TechSupport Forum}\footnote{\url{http://www.techsupportforum.comforums/}}, which support the unfolding of informal exchanges in the form of debates or discussions, amongst several users. It is acknowledged (e.g., in \cite{socialarg}) that computational argumentation
could benefit these online systems by supporting a formal analysis of the exchanges taking place therein~\cite{facebook}.

As far as we know, no in-depth study has already been accomplished on describing the specific small-world, or, more in general, network properties of interaction graphs in Argumentation. As a result, in ConArg we support the generation of small-world graphs according to two generic well-know kinds of properties (i.e., \emph{Barabasi} and \emph{Kleinberg}), and we leave the suggested elaboration to future work.

\section{Performance Tests}
\label{sec:tests}
The main goal of this section is to test ConArg.  All the following experiments are commented together at the end of this section, in order to give a panoramic view over them. To the best of our knowledge, these tests represent the first attempt  to find and tests Argumentation extensions in small-world networks.

To solve all the following problems we adopt a \emph{Depth First Search} (\emph{DFS}) algorithm~\cite{bookrossi}: this algorithm
 searches for a possible solution by organising the search space as a search tree. In every node of this tree a 
 value is assigned to a domain variable and a decision whether the node will be extended or the search will 
 be cut in this node is made. The search is cut if the assignment to the selected domain variable does not fulfil 
 all constraints. Each time during the search, we select the variable that has most constraints
assigned to it, and we assign to it a random value from its current domain: we use \emph{MostConstrainedStatic} as the variable selection heuristic and \emph{IndomainSimpleRandom} as the value selection heuristic, both natively offered by JaCoP. Using  the MostConstrainedStatic heuristic means that, since we test the tool with small-world/scale-free networks, we first select the hub nodes of the graph during the search: nodes with more links are inspected before the other ones.
Moreover, we set a timeout of $180$ seconds to interrupt the search
procedure and to report the number of solutions found only within that
time threshold.

\begin{figure}
  \centering
    \includegraphics[scale=0.4]{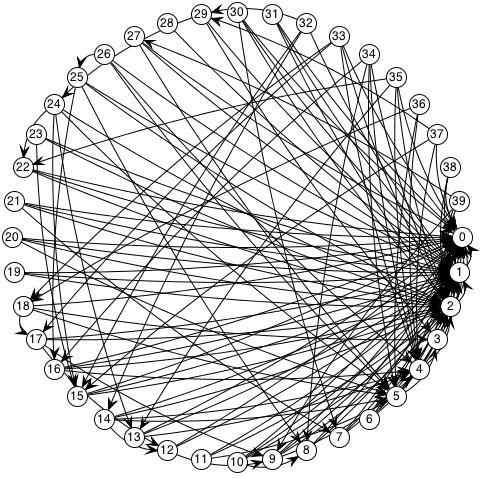}
    \caption{A small-world network with $40$ nodes, generated with JUNG by using the \emph{BarabasiAlbertGenerator} class~\cite{JUNG,Barabasi}. The big hubs that lead to  the small-world property are mainly nodes $0$, $1$ and $2$.}
    \label{fig:smallworld}
\end{figure}

For the first round of experiments
we use \emph{Barabasi} networks, whose properties are explained in item \emph{a)} of Section~\ref{ConArg}.
An example of such random graphs with $40$ nodes is shown in Figure~\ref{fig:smallworld}. These results are shown in Table~\ref{testscale}, and they are averaged over $10$ different random networks with respectively $10$, $20$, $30$, $32$, $37$, $40$, $60$ and $100$ nodes/arguments each. When the problem implies an exhaustive search we report the number of found extensions (i.e., conflict-free, admissible, complete and stable). In parentheses we also show the time (in milliseconds) needed to complete the search; when at least one of the $10$ random instances for each class exceeds the time threshold of $180$ seconds, we highlight this by using $\ast$ within parentheses. ``Grounded'' column in Table~\ref{testscale} only reports the number of milliseconds used to find the single solution, while ``Check if preferred'' column shows the time needed to check if a candidate extension is preferred or not (results are average over $20$ candidate extensions for each of the $10$ instances).

\begin{table}
\centering
\footnotesize{
  \begin{tabular}{| c | c | c | c |}
    \hline
    $\#$Nodes ($\#$edges) & $\#$Conf-free (ms) & $\#$Admissible (ms) & $\#$Complete (ms)\\ \hline
    10 (25) & 54 (2.73) & 26 (1.06) & 1 (0.02) \\ \hline
    20 (55) & 5,081 (283) & 498 (22.3) & 1 ($\sim$1) \\ \hline
    30 (75) & 385,176 (30,251) & 90,202 (7,113) & 1 ($\sim$1) \\ \hline
    32 (82) &  984,449 (85,677) & 105,392 (8,522) & 1 ($\sim$1) \\ \hline
    37 (82) &  2,233,186 ($\ast$) & 295,884 (28,256) & 1 ($\sim$1) \\ \hline
    40 (150) & 1,875,801 ($\ast$) & 933,782 ($\ast$) & 1 ($\sim$1) \\ \hline
    60 (250) & 1,303,049 ($\ast$) & 1,342,319 ($\ast$) & 1 ($\sim$1)\\ \hline
    100 (450) & 739,086 ($\ast$) & 698,084 ($\ast$) & 1 (1.71)\\ \hline
  \end{tabular} 
    \begin{tabular}{| c | c | c | c |}  
    \hline
    $\#$Nodes/$\#$edges & $\#$Stable (ms) & Grounded & Check if preferred \\ \hline
     10 (25) & 1 (0.03)&  1.3ms &  0.31ms \\ \hline
     20 (55) & 1 ($\sim$1) &  $\sim$1ms & $\sim$1ms \\ \hline
     30 (75) & 1 ($\sim$1) & $\sim$1ms & $\sim$1ms  \\ \hline
     32 (82) & 1 ($\sim$1) & $\sim$2ms & $\sim$1ms  \\ \hline
     37 (82) &1 ($\sim$1) & $\sim$2ms & $\sim$1ms  \\ \hline
     40 (150) & 1 ($\sim$1) & $\sim$3ms & $\sim$1ms  \\ \hline
     60 (250) & 1 ($\sim$1) & $\sim$3ms & $\sim$1ms  \\ \hline
     100 (450) & 1 (1.52) & 4.3ms  & 0.97ms \\ \hline
  \end{tabular} }
 \caption{We show the tests on eight different \emph{Barabasi} networks with respectively $10$, $20$, $30$, $32$, $37$, $40$, $60$ and $100$ nodes/arguments. $\#$ tags identify the number of elements (e.g., nodes or extensions). In parentheses we also report the number of milliseconds needed to find all the solutions; the $\ast$ tag means that the search for some of the $10$ random instances has been interrupted after the predefined threshold of $3$ minutes. ``Grounded'' and ``Check if preferred'' columns clearly report the time to solve the problem only.} \label{testscale} 
\end{table}



\begin{figure}
  \centering
    \includegraphics[scale=0.4]{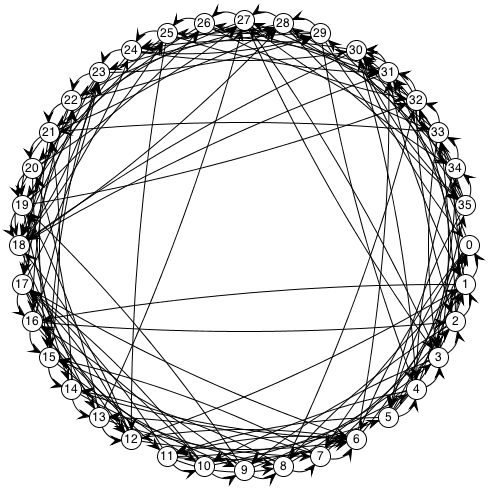}
    \caption{A small-world network with $36$ nodes generated with JUNG by using the \emph{KleinbergSmallWorldGenerator} class~\cite{JUNG,Barabasi}. Differently from the network in Figure~\ref{fig:smallworld}, here the small-world property is achieved through a two-dimensional grid structure and few long-distance links between nodes. No big hubs are present.}
    \label{fig:klein}
\end{figure}

In order to study our implementation on different networks,
we have repeated the same tests \emph{Kleinberg} networks, explained in item \emph{b)} of Section~\ref{ConArg}. We set a clustering coefficient of $0.5$ for all the tests over this kind of network.  An example of such graphs is shown in Figure~\ref{fig:klein}. In Table~\ref{testscale3} we report the performance collected with the same methodology as for Table~\ref{testscale}. The ``Grounded'' and ``Check if preferred'' columns are not reported in Table~\ref{testscale3}, since the obtained performance are similar to Table~\ref{testscale}.

\begin{table}
\centering
 \footnotesize{
  \begin{tabular}{| c | c | c | c | c |}
 
    \hline
    $\#$Nodes(edges) & $\#$Conf-free (ms) & $\#$Adm. (ms) & $\#$Compl. (ms) & $\#$Stable (ms) \\ \hline
    9 (45) &  21 (4.25) &  13 (5.58) & 10 (4.03) & 7 (1.54)\\ \hline
    25 (125) & 6,985 (595)  & 1,956 (195) & 533 (91) & 82 (12) \\ \hline
    36 (180) & 354,513 (36,296) & 63,560 (7,269) & 10,856 (1,966)& 541 (80)\\ \hline
    49 (245) & 1,418,333 ($\ast$) & 1,163,836 ($\ast$) & 273,330 (76,863)& 5,370 (1,151)\\ \hline
    64  (320) & 1,020,483 ($\ast$) & 931,105 ($\ast$) & 687,358 ($\ast$) & 73,315  (19,019)\\ \hline
    100 (500)  & 618,484 ($\ast$) & 591,537 ($\ast$)& 495,050 ($\ast$)& 515,615 ($\ast$)\\ \hline 
  \end{tabular} 
%
}
 \caption{We show the tests on six different \emph{Kleinberg} networks with respectively $9$, $25$, $36$, $49$, $64$  and $100$ nodes/arguments (as a remind, these networks have $n\times n$ nodes). The meaning of $\#$ and $\ast$ tags is the same as in Table~\ref{testscale}.} \label{testscale3} 
\end{table}

\begin{figure}
\centering
\includegraphics[scale=0.43]{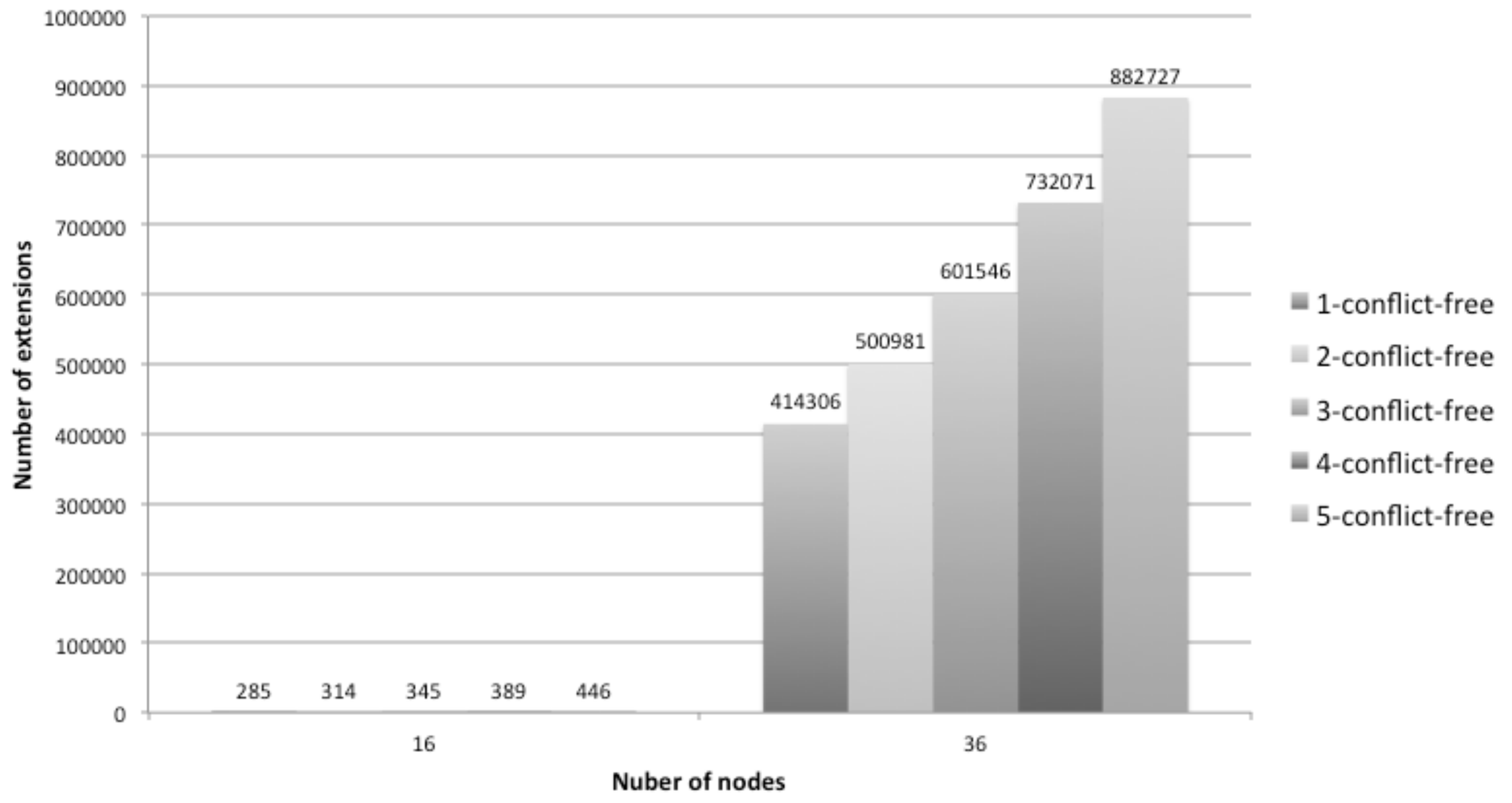} 
\caption{We show the number of $1$ (from left to right) up to $5$-conflict-free extensions found in \emph{Kleinberg}  networks with $16$ (grouped on the left) and $36$ (grouped on the right) nodes. Results are averaged on $10$ different networks each.} \label{figure:alphac}
\end{figure}

In the successive experiments we test how ConArg behaves over WAFs, that is argumentation frameworks with weights labelling the attacks (see Section~\ref{sec:bgweightedAF}). We have executed some tests concerning $\alpha$-conflict-free extensions (see Section~\ref{sec:argsemi}). The results are show in Figure~\ref{figure:alphac}, and they report the number of $1$ up to $5$-conflict-free extensions for \emph{Kleinberg} networks of $16$ and $36$ nodes.

\begin{figure}
\centering
\includegraphics[scale=0.45]{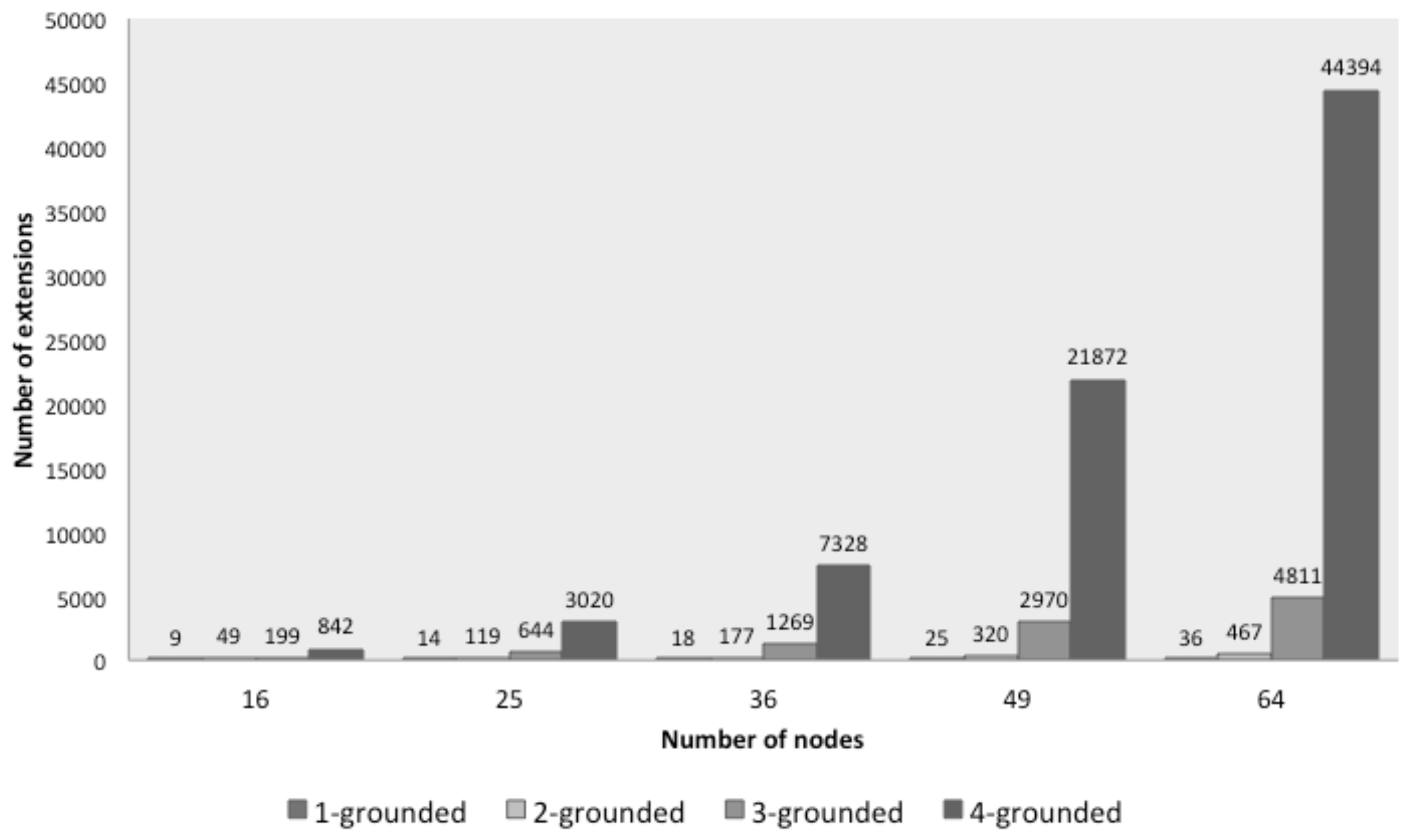} 
\caption{Number of $\beta$-grounded extensions~\cite{main,main2} in \emph{Kleinberg} networks of $16$, $25$, $36$, $49$ and $64$ nodes respectively. For each size, tests are executed with $\beta=1$, $\beta=2$, $\beta=3$ and $\beta=4$, and averaged over $10$ different random networks.} \label{figure:betaground}
\end{figure}

In addition, we show how the number of $\beta$-grounded extensions (see Section~\ref{sec:bgweightedAF}) scales while increasing the $\beta$ consistency budget. The results are reported in Figure~\ref{figure:betaground} for \emph{Kleinberg} networks of $16$, $25$, $36$, $49$ and $64$ nodes respectively; for each of these sizes, on \emph{y} axis we count the number of $\beta$-grounded extensions when $\beta$ is equal to $1$ up to $4$. Each of these results is averaged on $10$ different random networks.


In the following we list the global conclusions we collect from this section on performance: 
\begin{itemize}
\item As a first remark, we notice that the number of Dung's extensions strongly depends on the  topology of the considered interaction graph, even if these networks show the same small-world phenomenon. In particular, the most apparent feature of Barabasi networks is that they always show one complete and one stable extensions (which coincide), whatever the network size is (see Table~\ref{testscale}). Moreover, they always show a high number of conflict-free and admissible extensions, which grows very quickly with the number of nodes. On the contrary, Table~\ref{testscale3} shows that Kleinberg networks are ``more balanced'' in this sense, since we can find less conflict-free and admissible extensions, and up to hundreds of thousands complete and stable extensions. This is the reason why we think a deep study on the features of real argumentation networks is really important: their differences sensitively impact on the feasibility of working with them in an effective way.
\item The second issue concerns the feasibility of working with argumentation networks itself. From Table~\ref{testscale} and Table~\ref{testscale3} we can see that the number of conflict-free and admissible extensions explodes between $32$-$40$ nodes for both Barabasi and Kleinberg networks. Admissible extensions explode after $37$ nodes (see Table~\ref{testscale}). It is still possible to easily work with networks of $49$ and $64$ nodes, considering complete and stable extension respectively (see Table~\ref{testscale3}). These are the ``attention thresholds'' that should be taken into account when working with such networks. \item Constraint Programming performs extremely well on ``yes/no'' argumentation problems. For instance, checking if an extension is preferred is always solved almost instantaneously (see Table~\ref{testscale}), even if, as a remind, it is co-NP-complete problem~\cite{argcomputa}. However, as one could expect, Constraint Programming performs worse when it deals with the exhaustive enumeration of all the possible solutions, especially when the problem is loosely constrained, as for conflict-free extensions. Admissible, complete and stable extension represent a progressive refinement of conflict-free ones, through the addition of further constraints. In case of less constrained problems, propagation techniques are less effective, and the search space is consequently wider. Even if any complete search-method has to face this sudden state explosion, we are confident that improving the search with additional (maybe ad-hoc) heuristics can lead to better performance: for instance, we can detect and remove symmetries (Chapter $10$ in \cite{bookrossi}), or add global constraints (Chapter $6$ in \cite{bookrossi}) related the structure of the network. Note that these (and other possible) improvements strongly depends on the topology of small-world networks.
\item The presence of weights, that is in case of WAF, brings  more performance degradation when the goal is to enumerate all the solutions. The reason is that a certain amount of conflict is tolerated (see Section~\ref{sec:bgweightedAF} and Section.~\ref{sec:argsemi}), so that more solutions satisfy the relaxed problem. As we can see in Figure~\ref{figure:alphac} and Figure~\ref{figure:betaground}, the number of weighted extensions quickly increases as we allow for more tolerance: for instance, $4$-grounded extensions in networks with $64$ nodes (see Figure~\ref{figure:betaground}) are more than $922\%$ of the corresponding$3$-grounded extensions. Even larger proportions hold between $3$-grounded and $2$-grounded, and $2$-grounded and $1$-grounded extensions. We can suppose than anytime we increase the tolerance threshold by one, the number of extensions augments by one order of magnitude, barely for all the cases in Figure~\ref{figure:betaground}.  Figure~\ref{figure:alphac} shows that $\alpha$-conflict-free extensions (see Section.~\ref{sec:argsemi}) rapidly increase in large networks, since, for instance, $1$-conflict-free extensions are three order of magnitude more in networks with $36$ nodes than in networks with $16$ nodes. Their number increases less, but still considerably in networks with $36$ nodes, if the tolerance threshold is raised (e.g., $1$ to $2$-conflict-free): around $1000$ more extensions for every threshold increase of one unit (see Figure~\ref{figure:alphac}).
\end{itemize}


All the performance have been collected using a MacBook with a 2.4Ghz Core Duo processor and 4Gb 1067Mhz DDR3 of RAM.

\subsection{Comparison with ASPARTIX~\cite{aspartix}}\label{sec:aspartix}
The ASPARTIX tool\footnote{\url{www.dbai.tuwien.ac.at/proj/argumentation/systempage/}} ~\cite{aspartix,aspartix2}, which is based on \emph{Answer Set Programming} (\emph{ASP}), can be considered as the most complete and advanced system in literature for solving AFs and WAFs. ASPARTIX can be used not only  to compute the standard extensions for classical argumentation frameworks defined by Dung~\cite{dung}, but also for preference-based AF's (PAF's)~\cite{inconsistency}, value-based AF's (VAF's)\cite{vaf} and bipolar AF's (BAF's)~\cite{biparg}. In the latter case it is also possible to compute, save, and complete extensions, as well as distinguish between the classical d-admissible (following Dung), s-admissible (for stable) and c-admissible (for closed) extensions, for which also the respective preferred extensions are available. Furthermore, ASPARTIX is able to provide encodings for semi-stable and ideal semantics.

In order to execute it, it is required to use an ASP solver like \emph{Gringo/Clasp(D)}\footnote{\url{http://potassco.sourceforge.net}} or \emph{DLV}\footnote{\url{http://www.dlvsystem.com/dlvsystem/index.php/Home}}. Recent advances in ASP systems, in particular, the \emph{metasp optimization frontend} for the ASP-package Gringo/ClaspD provides direct commands to filter answer sets satisfying certain subset-minimality (or -maximality) constraints~\cite{aspartix2}. Since we decided to compare the two tools by considering only admissible, complete and stable extensions, we opted for the DLV system, because we do not need any minimality/maximality optimisation on the considered classes of problems.

We decided to compare ASPARTIX and ConArg on three different problems with the same (averaged on $10$) random Kleinberg networks: \emph{i)} finding all admissible extensions using $36$ nodes, \emph{ii)} finding all  complete extensions using $49$ nodes, and \emph{iii)} finding all  stable extensions using $64$ nodes. We have chosen these problems because, as we can notice in Figure~\ref{fig:klein}, the are computationally demanding, but still solvable within the threshold of $3$ minutes in ConArg.

\begin{figure}
\centering
\includegraphics[scale=0.43]{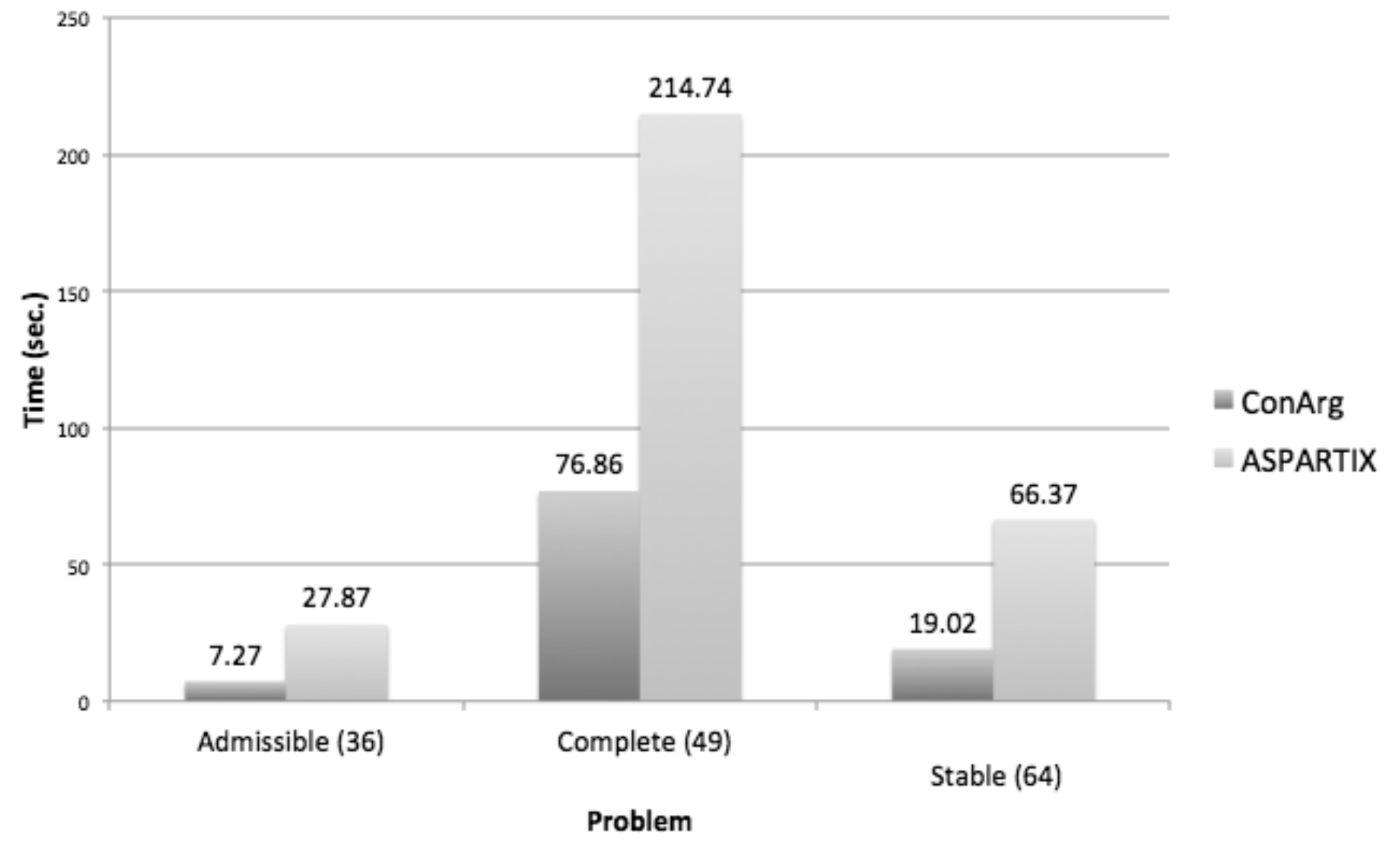} 
\caption{A time performance comparison between ASPARTIX~\cite{aspartix} and ConArg over three different problems: from left to right, finding all admissible extensions (in $10$ networks of $36$ nodes), finding all complete extensions (in $10$ networks of $49$ nodes), and finding all stable extensions (in $10$ networks of $64$ nodes). On the vertical axis we report the time (in seconds) needed to solve the problems.} \label{figure:comparison}
\end{figure}

In Figure~\ref{figure:comparison} we compare the different execution times of these problems, for both ASPARTIX and ConArg. To measure the time of ASPARTIX, we have used the \emph{OS X} terminal command ``\emph{time}''. We have summed \emph{User} and \emph{Sys} times: \emph{User} is the amount of CPU time spent in user-mode code (outside the kernel) within the process. \emph{Sys} is the amount of CPU time spent in the kernel within the process.  As for all the other tests, performance have been collected using a MacBook with a 2.4Ghz Core Duo processor and 4Gb 1067Mhz DDR3 of RAM. As we can se from the bars in Figure~\ref{figure:comparison}, ConArg outperforms ASPARTIX on all the three proposed problems. Performance in time are improved by respectively \emph{i)} $74\%$, \emph{ii)} $65\%$, and \emph{iii)} $72\%$.

\section{Related Work}
\label{sec:related} 
As far as we know, few systems have been proposed in literature to study AFs and (especially) WAFs from the computational point of view. To the best of our knowledge, the results presented in \cite{ictai11,tafa11} are among the first ones proposed on large problems, and the first ones using random networks showing small-world properties. By using ASPARTIX~\cite{aspartix}, the only other tests have been proposed in \cite{aspartix2}, where graphs  ranging from $20$ to $110$ arguments are  randomly generated. Two methods are used: the first generates arbitrary AFs and inserts for any pair $(a, b)$
the attack from $a$ to $b$ with a given probability $p$. The other method generates AFs with a $n \times m$ grid structure.  The tested extensions are the preferred, semi-stable, stage and resolution-based grounded semantics~\cite{aspartix2}. However, in this paper we have opted for testing our tool on small-world networks, since, in general, they show to be the most appropriate topology to represent social networks (see Section~\ref{ConArg}). Moreover, we also propose tests on WAFs and hard problems presented in \cite{main,main2}.

For example, in \cite{main,main2}, one of the main inspiration sources of this work (at least  for what concerns WAFs),  no solving mechanism is proposed to solve the problems presented in the paper. The focus is rather in defining the computationally hard problems, and proposing the related complexity proofs.

In \cite{corgias} the authors present \emph{GORGIAS-C}, which is a system implementing a logic programming framework of argumentation that integrates together preference reasoning and constraint solving. The system computes answers to queries asked on a logic program with priorities
on rules, and domain constraints on variables. GORGIAS-C is implemented as a modular meta-interpreter for its logic programs on top of \emph{Logtalk}\footnote{\url{http://logtalk.org}} using \emph{SWI-Prolog}\footnote{\url{http://www.swi-prolog.org}} and its ``\emph{Constraint Logic Programming (Finite Domain)}'' library and has successfully been used with \emph{ECLIPSe}\footnote{\url{http://eclipseclp.org}} with \emph{CLP} over reals. No computational results on problems related to AFs have been yet presented for this tool. Moreover, the system proposed in \cite{corgias} appears to be a more general framework for reasoning on multi-agent system, while our solution is more focused on the computational point of view.

In \cite{argcomputa} the authors associates to each subset $S$  of arguments a formula in propositional logic; then, $S$ is an extension under a given semantics if and only if the formula is satisfiable (i.e., they solve the problem with SAT~\cite{SATCP}). An extensive survey of the difference between SAT and CP can be found in \cite{SATCP}: summarizing, CP is more expressive in the modelling phase: this allows to find more complex semantics (e.g., grounded or semi-stable~\cite{semistable} ones) and further user-defined constraints on classical semantics~\cite{constrainedarg}.  In addition, in CP the user has the possibility to inform the solver about problem specific information and then to appropriately tune it, while in SAT there is usually little room and need for this parametrization. The modeling in \cite{argcomputa}  does not include preferred, grounded or weighted extensions~\cite{ecai10,main,main2}; furthermore, the encoding presented in \cite{argcomputa} has no practical implementation and performance tests.

In a very recent paper~\cite{sum11}, the authors present how to encode AFs as CSPs. They  show how to represent preferences over arguments as a (partial or total) preorder. In this work we have decided to model quantitative preferences instead of qualitative ones, even if qualitative preferences can be clearly cast  in our semiring-based framework either. Moreover, while in \cite{csclp09} some of the authors of this paper model the preferences over arguments, in this paper we associate weights with attacks instead, as also proposed in~\cite{ecai10,main,main2}. In addition to~\cite{sum11}, we provide a practical implementation of the constraint modelling (i.e., the ConArg tool) and performance tests.

The tool\footnote{\url{http://heen.webfactional.com}} described in \cite{caminadatool} provides a demonstration of a
number of basic argumentation components that can be applied
in the context of multi-agent systems. These components
include algorithms for calculating argumentation semantics,
as well as for determining the justification status
of the arguments and providing explanation in the form of
formal discussion games. Thus, even in this case the problem is not challenged from the computational point of view.

The  \emph{ASPARTIX} system~\cite{aspartix,aspartix2} is a tool
for computing acceptable extensions for a broad range of
formalizations of Dung's AFs and
generalisations thereof, e.g., value-based AFs~\cite{vaf} or
preference-based~\cite{inconsistency}. \emph{ASPARTIX} relies on a
fixed disjunctive \emph{Datalog} program which takes an instance of an
argumentation framework as input, and uses the Answer-Set solver
DLV for computing the type of extension specified by the user.
However, \emph{ASPARTIX} is not able to solve weighted AFs, as well as other ASP
systems~\cite{asp2}. Since ASPARTIX appears to be the most complete and frequently updated tool among the others, we selected it to be compared against ConArg in Section~\ref{sec:aspartix}, where time performance are shown and commented for both systems.

Finally, in \cite{sac11} some of the authors of this paper  extend classical AFs~\cite{dung} in order to deal with coalitions of arguments. The initial set of arguments is partitioned into subsets, or coalitions. Each coalition represents a different \emph{line of thought}, but all the coalitions show the same property inherited by Dung, e.g., all the coalitions in the partition are admissible (or conflict-free, complete, stable). Even this kind of problems based on coalitions can be solved by ConArg.

\section{Conclusions and Future Work}
\label{sec:conclusions}

We have presented ConArg, a constraint-based  tool programmed in Java, which can solve several problems related to AFs and WAFs. In this way, we have proposed an unifying computational
framework for Argumentation problems, with strong mathematical foundations and efficient solving
heuristics. Thanks to AI-based techniques, ConArg is able to efficiently solve some computationally hard extensions, as, for instance, the problems presented in \cite{main,main2} related to $\beta$-grounded extensions. 

The inspiration behind this work has been to study AFs/WAFs from the computational point of view, by developing a general common framework, then implementing the related tool and, finally, studying the problem by exploring the difficulties in practically solving such instances.

In addition, a second goal has been to link AFs/WAFs to small-world networks (see Section~\ref{ConArg}): all the tests in Section~\ref{sec:tests} have been performed over two different kinds of small-world networks. As far as we know, this corresponds to the first attempt in this direction. A comparison with ASPARTIX (see Section~\ref{sec:aspartix}) shows that constraint solving techniques prove to be
able to efficiently deal with large-scale problems. Practical applications of our ConArg may consist, for
example, in automatically studying discussion fora~\cite{zeno} or social networks~\cite{facebook} in general, where arguments may be rated by users leading to the definition of WAFs, with different strength values associated with the attacks.

For the future we have many open issues. We would like to investigate the properties of interaction graphs, in order to reproduce the tests we have presented in this paper on real-world cases (not only  generated in a random way). Therefore, we would like to set up our AFs/WAFs from real social networks, using real data. Close to this topic, we would also like to study the topology of real AFs, in order to further improve the performance of the tool with ad-hoc heuristics depending on the topology of the adopted networks.

\section*{Acknowledgement}
We would like to thank Bruno Capezzali, Davide Diosono, Valerio Egidi, Luca Mancini, Fabio Mignogna and Francesco Vicino, who developed the first version of ConArg as the final project of the exam ``Constraint Systems'', during their Master studies in Computer Science, at the University of Perugia.

\bibliographystyle{elsarticle-num}  
\bibliography{bibliotsccp,tesi,argumentation}
\end{document}